\documentclass{article} 
\usepackage{iclr2026_conference, times} 
\usepackage[T1]{fontenc}
\usepackage{multicol}
\usepackage{amsmath}
\usepackage{multirow}
 \usepackage{graphicx}
 \usepackage{enumitem}
 
\usepackage{amsmath,amsfonts,bm}

\def\eqref#1{equation~\ref{#1}}

\def\1{\bm{1}}

\DeclareMathAlphabet{\mathsfit}{\encodingdefault}{\sfdefault}{m}{sl}
\SetMathAlphabet{\mathsfit}{bold}{\encodingdefault}{\sfdefault}{bx}{n}

\usepackage{url}
\usepackage{xcolor}
\usepackage{amssymb}
\usepackage{pifont}
\usepackage{caption} 
\usepackage{tabularx} 
\usepackage{CJKutf8}
\usepackage{adjustbox}
\usepackage{colortbl} 
\usepackage{xspace}
\usepackage{subcaption}
\usepackage{titlesec}

\usepackage{ragged2e}

\usepackage{caption}
\usepackage{xurl}
\usepackage{hyperref}

\usepackage{tabularx}
\usepackage{booktabs}
\usepackage[table,dvipsnames]{xcolor}

\titleformat{\paragraph}[runin]
  {\normalfont\normalsize\bfseries}
  {}
  {0em}
  {}
\titlespacing*{\paragraph}
  {0pt} 
  {0pt} 
  {1em} 
\titlespacing{\section}{0pt}{2ex}{1ex}
\titlespacing{\subsection}{0pt}{1.5ex}{0.8ex}
\newcommand{\Nmark}{\textcolor{red}{\ding{55}}}
\newcommand{\Ymark}{\textcolor{green}{\checkmark}}

\newcolumntype{P}[1]{>{\centering\arraybackslash}m{#1}}

\newcommand{\TextIcon}{\includegraphics[height=0.9em]{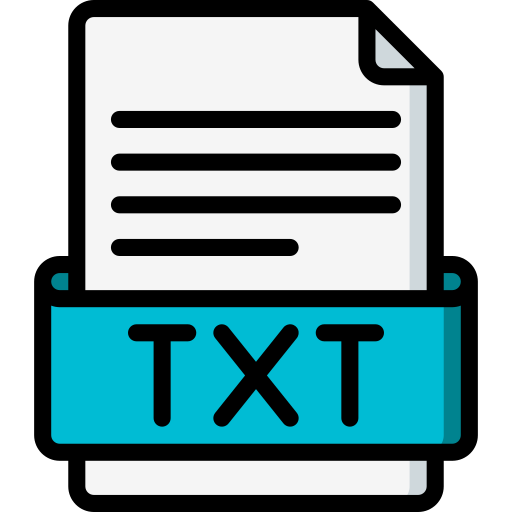}\xspace}
\newcommand{\ImageIcon}{\includegraphics[height=1em]{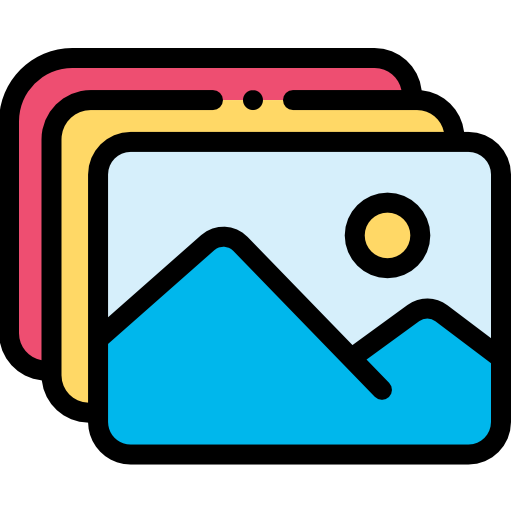}\xspace}
\newcommand{\SpeechIcon}{\includegraphics[height=1em]{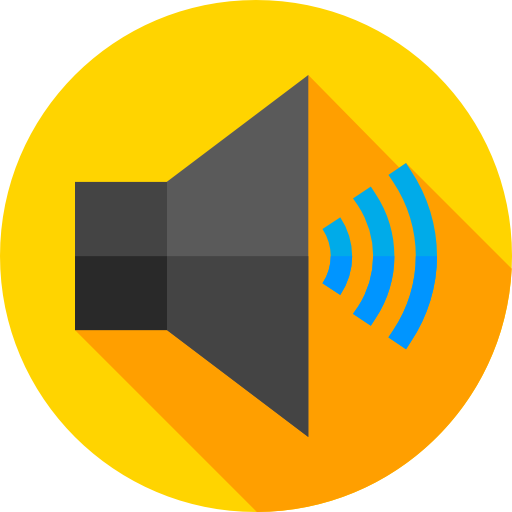}\xspace}
\newcommand{\BlueColorIcon}{\includegraphics[height=0.6em]{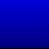}\xspace}
\newcommand{\RedColorIcon}{\includegraphics[height=0.6em]{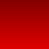}\xspace}

\newcolumntype{Y}{>{\RaggedRight\arraybackslash}X}
\newcolumntype{L}{>{\RaggedRight\arraybackslash}X}

\newsavebox{\myfirsttablebox}
\newsavebox{\mysecondtablebox}
\newlength{\maxtablewidth}

\title{MMA-ASIA: A Multilingual and Multimodal Alignment Framework for Culturally-Grounded Evaluation}

\author{%
  \parbox{\textwidth}{%
    \raggedright
    \vspace{0.8em}
    {\bfseries
    Weihua Zheng$^{1,2}$ \quad
    Zhengyuan Liu$^{2}$ \quad
    Tanmoy Chakraborty$^{3}$ \quad
    Weiwen Xu$^{4}$ \\
    Xiaoxue Gao$^{2}$ \quad
    Bryan Chen Zhengyu Tan$^{1}$ \quad
    Bowei Zou$^{2}$ \quad
    Chang Liu$^{6}$ \quad
    Yujia Hu$^{1}$ \\
    Xing Xie$^{5}$ \quad
    Xiaoyuan Yi$^{5}$ \quad
    Jing Yao$^{5}$ \quad
    Chaojun Wang$^{4}$ \quad
    Long Li$^{4}$ \quad
    Rui Liu$^{7}$ \\
    Huiyao Liu$^{7}$ \quad
    Koji Inoue$^{8}$ \quad
    Ryuichi Sumida$^{8}$ \quad
    Tatsuya Kawahara$^{8}$ \quad
    Fan Xu$^{9}$ \\
    Lingyu Ye$^{9}$ \quad
    Wei Tian$^{9}$ \quad
    Dongjun Kim$^{10}$ \quad
    Jimin Jung$^{10}$ \quad
    Jaehyung Seo$^{10}$ \\
    Nadya Yuki Wangsajaya$^{11}$ \quad
    Pham Minh Duc$^{2}$ \quad
    Ojasva Saxena$^{3}$ \quad
    Palash Nandi$^{3}$ \\
    Xiyan Tao$^{2}$ \quad
    Wiwik Karlina$^{2}$ \quad
    Tuan Luong$^{2}$ \quad
    Keertana Arun Vasan$^{11}$ \\
    Roy Ka-Wei Lee$^{1}$\thanks{Corresponding authors.} \quad
    Nancy F. Chen$^{2}$\footnotemark[1]
    }\\[0.6em]
    {\normalfont\small
    $^{1}$Singapore University of Technology and Design \\
    $^{2}$Agency for Science, Technology and Research, Singapore \\
    $^{3}$Indian Institute of Technology Delhi \quad
    $^{4}$Alibaba DAMO Academy \\
    $^{5}$Microsoft Research Asia \quad
    $^{6}$Shanghai University of Finance and Economics \\
    $^{7}$Inner Mongolia University \quad
    $^{8}$Kyoto University \quad
    $^{9}$Jiangxi Normal University \\
    $^{10}$Korea University \quad
    $^{11}$Nanyang Technological University
    }%
  }%
}

\iclrfinalcopy 
\begin{document}
\begin{CJK}{UTF8}{gbsn}

\maketitle

\begin{abstract}

Large language models (LLMs) are now used worldwide, yet their multimodal understanding and reasoning often degrade outside Western, high-resource settings. We propose MMA-ASIA, a comprehensive framework to evaluate LLMs’ cultural awareness with a focus on Asian contexts. MMA-ASIA centers on a human-curated, multilingual, and multimodally aligned multiple-choice benchmark covering 8 Asian countries and 10 languages, comprising 27,000 questions; over 79\% require multi-step reasoning grounded in cultural context, moving beyond simple memorization. To our knowledge, this is the first dataset aligned at the input level across three modalities: text, image (visual question answering), and speech. This enables direct tests of cross-modal transfer. Building on this benchmark, we propose a five-dimensional evaluation protocol that measures -- (i) cultural-awareness disparities across countries, (ii) cross-lingual consistency, (iii) cross-modal consistency, (iv) cultural knowledge generalization, and (v) grounding validity. To ensure rigorous assessment, a Cultural Awareness Grounding Validation Module detects ``shortcut learning'' by checking whether the requisite cultural knowledge supports correct answers. Finally, through comparative model analysis, attention tracing, and an innovative Vision-ablated Prefix Replay (VPR) method, we probe why models diverge across languages and modalities, offering actionable insights for building culturally reliable multimodal LLMs.

\end{abstract}

\section{Introduction}

Large language and vision–language models are being increasingly deployed across various cultures and languages. Yet, their behavior remains uneven; performance is strongest in high-resource, Western contexts and degrades in non-Western settings, particularly across Asia \citep{chiu2025culturalbenchrobustdiversechallenging,romero2024cvqaculturallydiversemultilingualvisual,vayani2025languagesmatterevaluatinglmms,wang2024seaevalmultilingualfoundationmodels,myung2025blendbenchmarkllmseveryday}. As multimodal, multilingual models proliferate \citep{chen2024internvlscalingvisionfoundation,bai2023qwenvlversatilevisionlanguagemodel,jiang2023mistral7b,openai2024gpt4ocard,touvron2023llamaopenefficientfoundation}, evaluating whether they hold \emph{consistent} cultural interpretations across languages and modalities is both important and underexplored. 

In this paper, we investigate (i) cultural awareness consistency, defined as the extent to which a model gives stable answers to semantically equivalent inputs when the representation (text, image+question, or spoken question) or the language changes; (ii) cultural awareness grounding, defined as whether correct answers rely on appropriate cultural signals rather than exploitable shortcuts; and (iii) cultural awareness generalization, defined as whether a model that has access to the relevant cultural knowledge can perform the required reasoning within those cultural contexts \cite{balepur-etal-2024-artifacts,molfese2025right,zheng2023large}.
Nevertheless, conducting such evaluations presents significant challenges. Existing culture-centric datasets (e.g., \cite{myung2025blendbenchmarkllmseveryday, wang2024seaevalmultilingualfoundationmodels}) frequently suffer from two key limitations: (i) insufficient alignment of instances across modalities, (ii) inadequate representation of low-resource Asian languages. Furthermore, evaluation processes are easily hacked through memorization or elimination in multiple choice questions (MCQs), which bypass the genuine reasoning capabilities \citep{wang2025generalizationvsmemorizationtracing, hartmann2023sokmemorizationgeneralpurposelarge}. As a result, we still lack a principled way to separate actual cultural competence from artifacts.

To address this research gap, we introduce \textbf{MMA-ASIA}\footnote{We will release the data, splits, prompts, decoding settings, and per-item metadata (e.g., knowledge points, reasoning tags) for benchmarking, reproducing, and future extensions.}, an explainable evaluation framework for Asian cultural knowledge. MMA-ASIA aligns tri-modal items (textual question, image+question, and Text-to-Speech (TTS)-spoken question) with identical semantics and provides parallel local-language and English versions authored by native experts across 8 countries and 10 languages to make a comprehensive evaluation. The framework measures five axes: (1) cultural awareness disparity, (2) cross-modal consistency, (3) cross-lingual consistency, (4) cultural knowledge generalization under held-out regimes, and (5) grounding validation via targeted ablations and negative controls. 

Using MMA-ASIA, we evaluate 15 multilingual and multimodal LLMs(e.g., GPT-4o, Qwen, Llama). We find that (i) accuracy drops markedly in low-resource Asian languages compared to English, (ii) cross-modal consistency lags text-only performance, indicating incomplete transfer from language to vision and speech, and (iii) grounding controls reduce a non-trivial fraction of apparent ``wins,'' revealing shortcut use. We also analyze multi-step, culture-specific reasoning errors and where visual or linguistic cues fail to connect.
We summarize our contributions as follows:

\begin{itemize}[leftmargin=*,topsep=4pt]
\setlength{\itemsep}{0pt}
\setlength{\parskip}{0pt}
\setlength{\parsep}{0pt}
    \item \textbf{Aligned tri-modal, multilingual benchmark.} We release \textbf{MMA-ASIA} with 27,000 multilingual multimodal questions authored by in-country experts across 8 countries and 10 languages.  
    \item \textbf{Five-axis evaluation protocol.} We formalize cultural awareness \emph{consistency} (cross-modal/cross-lingual) and \emph{grounding} with negative controls and ablations, plus generalization tests under held-out themes/countries. We provide reference implementations and CI-tested evaluation scripts.
    \item \textbf{Extensive baselines and analyses.} We report zero-shot baselines for 14 model families (multilingual LLMs and VLMs), including common-support subsets, and diagnose failure modes by modality, language, and reasoning step count.

\end{itemize}
\noindent

{ 
\begin{table}[t!]
\centering
\footnotesize 
\caption{\textbf{Comparison of existing culture-related benchmark datasets with MMA-Asia (Ours).} \TextIcon, \ImageIcon\ and \SpeechIcon represent the the text, image and speech modalities, respectively. ``MLA'' and ``MDA'' denote ``multilingual alignment'' and ``multimodal alignment'', respectively.}
\label{tab:cultural-benchmarks}
\setlength{\tabcolsep}{2pt}
\renewcommand{\arraystretch}{0.5}
\rowcolors{2}{white}{gray!10} 
\setlength{\abovecaptionskip}{2pt} 

\begin{tabular}{@{}%
P{2.4cm} P{0.8cm} P{0.8cm} P{0.8cm} P{0.8cm} P{0.8cm} P{1.4cm} P{1.4cm} P{1.2cm} P{1.2cm} P{1cm}         
@{}}
\hline
Benchmark & \shortstack{MLA} & \shortstack{MDA} & \shortstack{RI}
& \shortstack{\# of\\ Ctries} & \shortstack{\# of\\ Langs}
& Modality & \shortstack{Multi-step\\ reasoning} & Question forms & \shortstack{Total\\ samples} & \shortstack{Total \\domains} \\
\hline
SeaEval            & \Nmark & \Nmark & \Nmark & 4  & 1   & \TextIcon                & \Ymark & Diverse & 415      & - \\
CLIcK              & \Nmark & \Nmark & \Nmark & 1  & 2   & \TextIcon                & -     & Diverse & 1,995    & 11 \\
BLEnD              & \Ymark & \Nmark & \Nmark & 16 & 13  & \TextIcon                & -     & Fixed   & 52,557   & 6  \\
Culturalbench      & \Nmark & \Nmark & \Nmark & 45 & 1   & \TextIcon                & -     & Fixed   & 1,696    & 17 \\
CVQA               & \Ymark & \Nmark & \Nmark & 30 & 31  & \TextIcon, \ImageIcon         & -     & Fixed   & 10,000   & 10 \\
CulturalbenchVQA   & \Nmark & \Nmark & \Nmark & 11 & 1   & \TextIcon, \ImageIcon         & -     & Diverse & 2,378    & 5  \\
ALM-bench          & \Ymark & \Nmark & \Nmark & 73 & 100 & \TextIcon, \ImageIcon         & -     & Diverse & 22,763   & 19 \\
Md3                & \Nmark & \Nmark & \Nmark & 3  & 1   & \SpeechIcon              & -     & Diverse & 3,689    & 2  \\
MULTI-AUDIOJAIL    & \Ymark & \Nmark & \Nmark & 6  & 6   & \SpeechIcon              & -     & Diverse & 102,720  & - \\
\rowcolor{pink!15}
Ours               & \Ymark & \Ymark & \Ymark & 8  & 10  & \TextIcon, \ImageIcon, \SpeechIcon & \Ymark & Diverse & 27,000   & 9 \\
\hline
\end{tabular}
\vspace{-0.3cm}
\end{table}
}

\section{Related Work}

\paragraph{Cultural knowledge in text.}
Recent benchmarks assess culture-specific knowledge via MCQs \citep{kim-etal-2024-click,wang2024seaevalmultilingualfoundationmodels,susanto2025seahelmsoutheastasianholistic,myung2025blendbenchmarkllmseveryday,chiu2025culturalbenchrobustdiversechallenging}, consistently showing (i) performance gaps favoring English/high-resource settings and (ii) sensitivity to formatting. However, most lack aligned multimodal counterparts to test cross-modal cultural understanding.

\paragraph{Cultural perception in images (VQA).}
Multilingual VQA datasets use community-sourced images and questions \citep{romero2024cvqaculturallydiversemultilingualvisual,nayak2024benchmarkingvisionlanguagemodels,vayani2025languagesmatterevaluatinglmms}, revealing vision-language gaps and language sensitivity. Yet they typically lack text/speech-parallel versions of identical items, making it hard to isolate whether failures stem from cultural knowledge, visual grounding, or language handling.

\paragraph{Speech and accent robustness.}
Speech datasets reveal substantial accent-related biases \citep{eisenstein2023md3multidialectdatasetdialogues,roh2025multilingualmultiaccentjailbreakingaudio}, and perturbations can drastically change outcomes. However, they rarely evaluate cultural knowledge directly or analyze trimodal alignment and  consistency.

\paragraph{Consistency and grounding.}
MCQ performance may reflect shortcuts rather than grounded reasoning \citep{balepur-etal-2024-artifacts,molfese2025right,zheng2023large}. Benchmarks seldom include negative controls or report cross-lingual/cross-modal consistency—essential for distinguishing cultural competence from artifacts.
\paragraph{How is MMA-ASIA different from others?}
Existing datasets evaluate cultural understanding within a single modality or language at a time, without tightly \emph{aligning} instances across modalities and languages, and no built-in \emph{grounding controls}. \text{MMA-ASIA} addresses these gaps by: (i) providing semantically aligned tri-modal items (text, image+question, speech) in parallel local-language and English versions; (ii) adopting \emph{cross-modal} and \emph{cross-lingual} consistency as prime metrics; and (iii) integrating targeted ablations and negative controls to test whether answers rely on the intended cultural signal rather than shortcuts. This design enables clearer attribution of failure modes: knowledge vs. language vs. modality, and more reliable measurement of cultural awareness in multimodal, multilingual models. We summarize key differences among representative datasets in Table~\ref{tab:cultural-benchmarks}.

\begin{figure*}[t!]
  \centering
  \includegraphics[trim=0 58 0 0, clip,width=\textwidth]{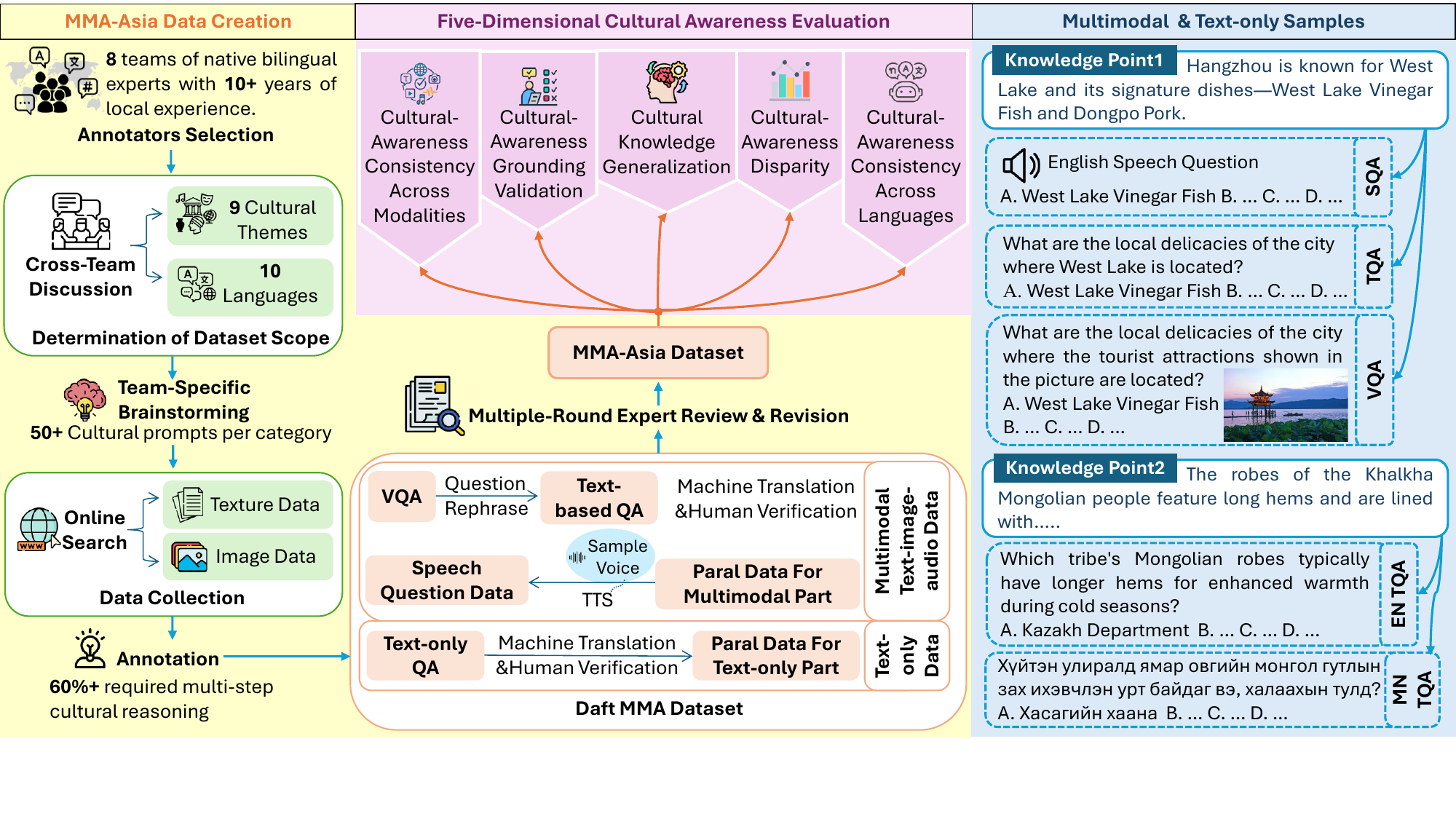}
  \caption{\textbf{An overview of the MMA-Asia evaluation framework:} data creation pipeline, representative dataset samples, and evaluation dimensions.}
  \label{fig: MMA flow Chart}
  \vspace{-0.5cm}
\end{figure*}

\section{Benchmark Construction}
\text{MMA-ASIA} was collaboratively constructed by research teams from eight countries: China, Singapore, Japan, South Korea, Mongolia, Vietnam, Indonesia, and India. The pipeline comprised five stages: (i) annotator selection, (ii) selection of representative cultural themes and languages, (iii) collection of text and image materials, (iv) question authoring and annotation by country sub-teams, and (v) human review and revision for quality and cultural representativeness. For the definition of cultural themes, we followed the framework proposed by \cite{adilazuarda-etal-2024-towards}.

\subsection{Annotator Selection}
An in-country expert team curated each national subset. All team members were native speakers of the local language and proficient in English. Annotators had lived in the respective country for more than ten years, ensuring deep familiarity with local cultural contexts. Detailed annotator information is in Appendix~\ref{sec:Annotator Demographic}. Before annotation began, we held project-wide briefings to explain the scope and requirements. We also distributed a detailed English annotation guideline (see Appendix \ref{sec:annotation guideline}).

\subsection{Cultural Themes and Languages}
Through collaborative discussions, we finalized 9 cultural themes, \emph{Daily life habits/Culture, Food/Cuisine, Transportation, Buildings, History, Geographical location and climate, Education, Fashion/Clothing and Language/Race}. We balanced the number of questions across themes as much as possible. Each national subset includes the country’s official language(s) and English. For India, we selected Hindi as the representative language due to its large speaker base among the 22 official languages. For Singapore, we included all four official languages: English, Chinese, Malay, and Tamil. In total, \text{MMA-ASIA} covers ten languages; full details are available in Appendix \ref{sec:data statistic}.

\subsection{Text and Image Data Collection}
For both the text-only and multimodal tracks, teams generated at least 56 keywords or short phrases per category as \emph{cultural prompts}. If a category could not supply enough prompts, the shortfall was filled using prompts from other categories. These prompts were designed to capture both diversity and geographic breadth, reducing the risk of homogenizing Asian cultures. Using these prompts, team members retrieved relevant texts and images from the web and extracted short passages to serve as the basis for question authoring. When source content was ambiguous, we cross-checked with multiple references to ensure authenticity. All images were obtained from Creative Commons (CC)-licensed resources (details in Appendix~\ref{sec:data statistic}).

\subsection{Question Creation and Annotation}
Teams used collected materials to create multi-choice QA data, requiring at least 60\% of questions to involve multi-step cultural reasoning. Multi-step reasoning questions require sequential derivation and/or synthesis from multiple independent knowledge components, not just single-fact recall or paraphrase (detailed examples in Appendix \ref{sec:data statistic}). Question templates are not fixed for preserving variety in question styles. Each national subset has two components: a \emph{multimodal} and a \emph{text-only} component. All QA data were authored in official local languages and translated to English using Claude 4 for Tamil and GPT-4o for other languages.
All translations underwent manual verification; mistranslations were corrected, and for terms lacking standard English equivalents, we applied phonetic transliteration or adopted the locally prevalent rendering.

\vspace{4pt}
\textbf{Multimodal Component.}
Annotators created VQA items where the correct answer requires visual understanding. Each VQA question was rephrased into a semantically-equivalent text-only MCQ. The answer options and the correct answer were kept unchanged. We provided both the original and rephrased items in English and the local language. We also generated speech inputs by converting the text to audio using high-quality TTS systems (Appendix \ref{sec:TTS tools details} provides TTS toolkit and speech data building details). For Spoken QA, we considered two configurations: (i) converting only the question stem to speech while keeping textual options, and (ii) converting both the stem and the options to speech. To preserve comparability with VQA under controlled variables and to reduce ambiguity introduced by fully spoken options, our main experiments adopt the ``spoken stem + textual options'' configuration across five evaluation dimensions. Results for the fully spoken setting (spoken stem and options) on the test set are reported in Appendix \ref{sec:results for fully spoken} for reference.
 To reflect accent effects, we produced English speech in both accent-neutral and locally accented versions.

\vspace{4pt}
\textbf{Text-only Component.}
This component contains questions that were not suitable for pairing with an image or are inherently text-based. We applied the same requirement that at least 60\% of questions involve multi-step reasoning. All questions were created in multiple languages.

For each question, annotators additionally identified the requisite \emph{knowledge points}. As shown in Fig. \ref{fig: MMA flow Chart}, these denote the minimal information necessary to arrive at the correct answer, usually summarized in a few concise sentences. These knowledge points are included with the dataset and support the evaluation of whether model explanations reflect culturally authentic reasoning.

For each question, annotators identified the requisite knowledge points, the minimal information needed for the correct answer, typically summarized in a few sentences (shown in Fig. \ref{fig: MMA flow Chart}). These knowledge points are used for model's cultural awareness grounding validation (Section \ref{sec:Cultural Awareness Grounding Validation}).

\subsection{Human Review and Revision}
After each country team completed a draft, in-country linguists conducted quality reviews. The review covered: ambiguity in wording, accuracy of English translations, clarity and fluency of the speech data, completeness of knowledge points, and appropriateness of answer options. Teams revised their subsets based on this feedback, yielding the final high-quality release.

\section{Experiments}
\label{sec:All_exp}
We evaluate existing LLMs on the \text{MMA-ASIA} benchmark. Unless stated otherwise, all runs are \emph{zero-shot} with a unified prompt template whose language matches the question language (prompts and experimental settings are in Appendix \ref{sec:testing prompt}). We report results along five dimensions and, for each, analyze the factors that drive performance. We access three closed-source models: GPT-4o~\citep{openai2024gpt4ocard}, Claude-Sonnet-4, and Gemini~2.5~Pro~\citep{comanici2025gemini25pushingfrontier}, and eleven open-source multilingual or multimodal models, including the Qwen~\citep{bai2025qwen25vltechnicalreport,qwen3technicalreport}, LLaMA~\citep{touvron2023llamaopenefficientfoundation}, and GLM~\citep{glm2024chatglm,vteam2025glm45vglm41vthinkingversatilemultimodal} families. Models and tasks are detailed in Appendix~\ref{sec:Model Selection}. For models without multilingual support, we report English-only scores for comparability. For speech evaluation, we include only models that accept \emph{speech tokens} directly; models that require intermediate automatic speech recognition (ASR) are excluded.

\begin{figure}[t!]
  \centering

  \begin{subfigure}{\textwidth}
    \centering
    \includegraphics[trim=0 30 0 10, width=\linewidth]{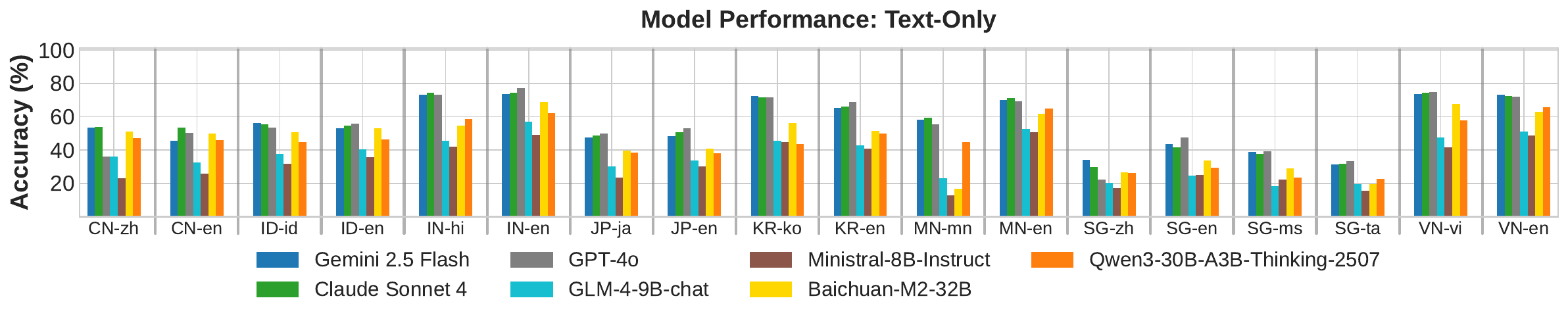}
    \label{fig:sub1}
  \end{subfigure}


  \begin{subfigure}{\textwidth}
    \centering
    \includegraphics[trim=0 64 0 10,width=\linewidth]{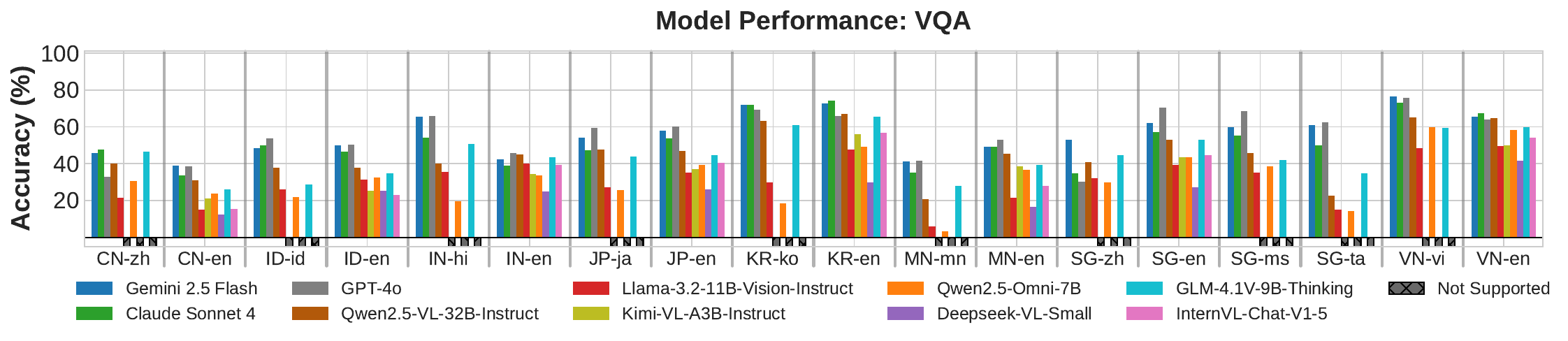}
    \label{fig:sub2}
  \end{subfigure}


  \begin{subfigure}{\textwidth}
    \centering
    \includegraphics[trim=0 30 0 10,width=\linewidth]{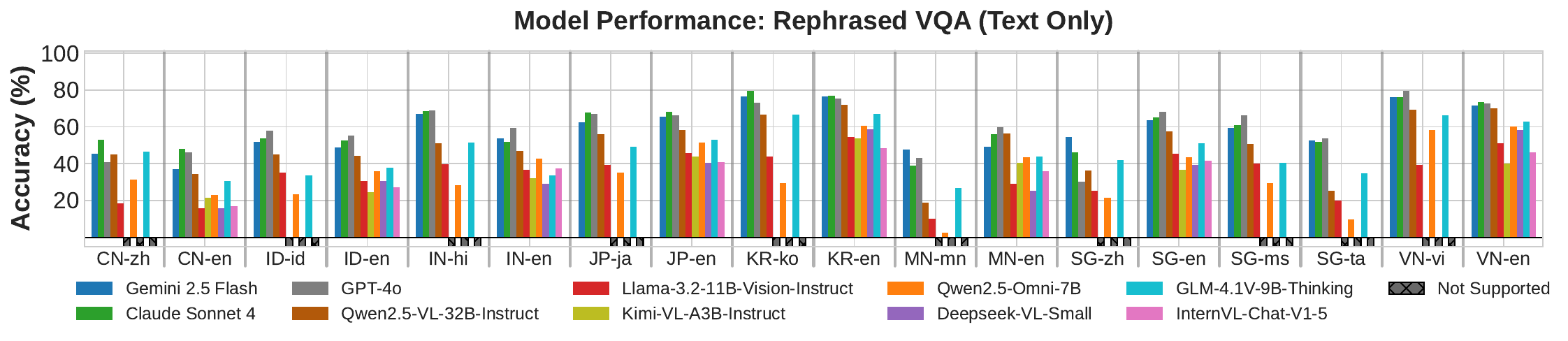}
    \label{fig:sub3}
  \end{subfigure}

  \begin{subfigure}{\textwidth}
    \centering
    \includegraphics[trim=0 40 0 10,width=\linewidth]{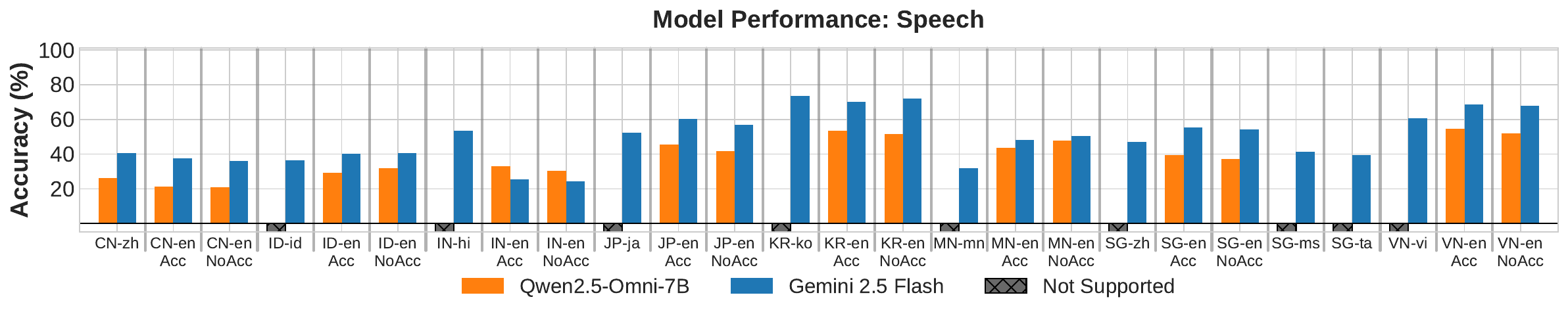}
    \label{fig:sub4}
  \end{subfigure}
  \caption{\textbf{Performance of LLMs on MMA-Asia across modalities.} Exact values are provided in Appendix \ref{sec:Performance of LLMs}. For each country and modality, the dataset contains 500 questions presented in multiple languages. The vertical axis reports Accuracy (\%), defined as the number of items where the model’s chosen option exactly matches the correct option, divided by 500. The x-axis label \emph{\{Country\}}--\emph{\{Language\}} denotes the cultural dataset for \emph{\{Country\}}, presented in \emph{\{Language\}}.}
  \label{fig:all}
  \vspace{-0.7cm}
\end{figure}

\subsection{Main Results}
Figure~\ref{fig:all} summarizes accuracy on \text{MMA-ASIA}. Nearly all state-of-the-art models score below 80\% and most below 50\%, highlighting the benchmark’s difficulty. Closed-source models outperform open-source models; even the strongest open-source family (Qwen) trails the closed-source average by more than 10 percentage points. Performance varies with (i) the resource level of the language and (ii) the evaluation modality. The following subsections analyze: (1) cultural-awareness disparities across countries and languages, (2) cross-lingual and cross-modal consistency, (3) cultural-awareness grounding, and (4) cultural knowledge generalization.

\subsection{cultural Awareness Disparity}
\label{sec:cultural Awareness Disparity}

\paragraph{Across countries (language factor controlled).}
To isolate country effects, we compare each model’s scores across countries within a fixed modality and, for each country, retain the model’s \emph{best} score over its available languages (Figure~\ref{fig:all}, Tables~\ref{tab:text-only},~\ref{tab:VQA},~\ref{tab:rephrase_vqa},~\ref{tab:speech_question_only}). For example, in the text-only setting, GPT-4o scores 71.4\% in Korean and 68.8\% in English; we use the precision in English to represent the awareness of Korean culture in GPT-4o, to avoid confounding cultural competence with language proficiency. Models show higher awareness for Korean culture on average (63.98\% across 4 modalities), plausibly due to global diffusion and richer data availability~\cite{jang2024more,dal2018analysis}. Remarkably, Vietnamese culture (62.96\%) is on par with Korean, likely reflecting Vietnam’s high social-media penetration (79.8\% of the population) and thus large volumes of user-generated content~\citep{DataReportalVietnam2025}. In contrast, China and India exhibit larger gaps, consistent with multilayered cultural forms and greater regional heterogeneity. Mongolia trails further, consistent with low-resource language settings and sparser training corpora.

\vspace{4pt}
\paragraph{Across languages (within culture).}
English prompts often outperform low-resource languages, reflecting the breadth of English corpora and limited cross-lingual transfer~\citep{hu2025largelanguagemodelscrosslingual,zheng2025cclxcotefficientcrosslingualknowledge,zheng2025adamcotrethinkingcrosslingualfactual}. This advantage diminishes or reverses for medium/high-resource local languages (e.g., Chinese, Japanese), where culture-specific terms and proper names are well represented locally but rare in English corpora, hurting retrieval and grounding. For example, “乌护 (Wuhu),” a lineage among Uyghur ancestral groups, lacks a standard English equivalent; transliteration is rare and ambiguous in English data. Thus, when the model is competent in the relevant local language, using that language can yield better cultural grounding than English.

\vspace{4pt}
\paragraph{Across modalities (holding language fixed).}
We observe a consistent ordering: \emph{text-only} $>$ \emph{VQA} $>$ \emph{spoken QA}. Data availability follows the same order (text $\gg$ image–text $\gg$ raw speech). Speech adds uncertainty (noise, homophony), and many architectures encode modalities separately and then fuse downstream, introducing alignment/compression losses that widen gaps. Interestingly, in Speech, Qwen and Gemini outperform their standard English baselines in 6 and 5 country-specific cultural settings, respectively (Figure~\ref{fig:all}). Accents appear to serve as a prior cue for specific cultures, enhancing the models' accuracy on corresponding tasks. We attribute this to the co-occurrence of accents and their related cultural content within the data (see Appendix~\ref{sec: Speech as a Cultural Prior} for detailed analysis).

From the foregoing analysis, it is evident that LLMs in Asian cultural contexts also display cultural and modality biases shaped by data distributions; furthermore, given the limited effectiveness of cross-lingual cultural knowledge transfer, English cannot be assumed to perform reliably better on culture-related tasks. In contrast, in the speech modality, accents, often treated as noise, paradoxically serve as effective cultural cues that activate relevant context and improve performance. In speech, accents, typically considered noise, actually serve as effective cultural cues that activate relevant context and improve performance.

\begin{figure}[t!]
  \centering
  \begin{subfigure}{\textwidth}
    \centering
    \includegraphics[trim=0 5 0 20, clip,width=\linewidth]{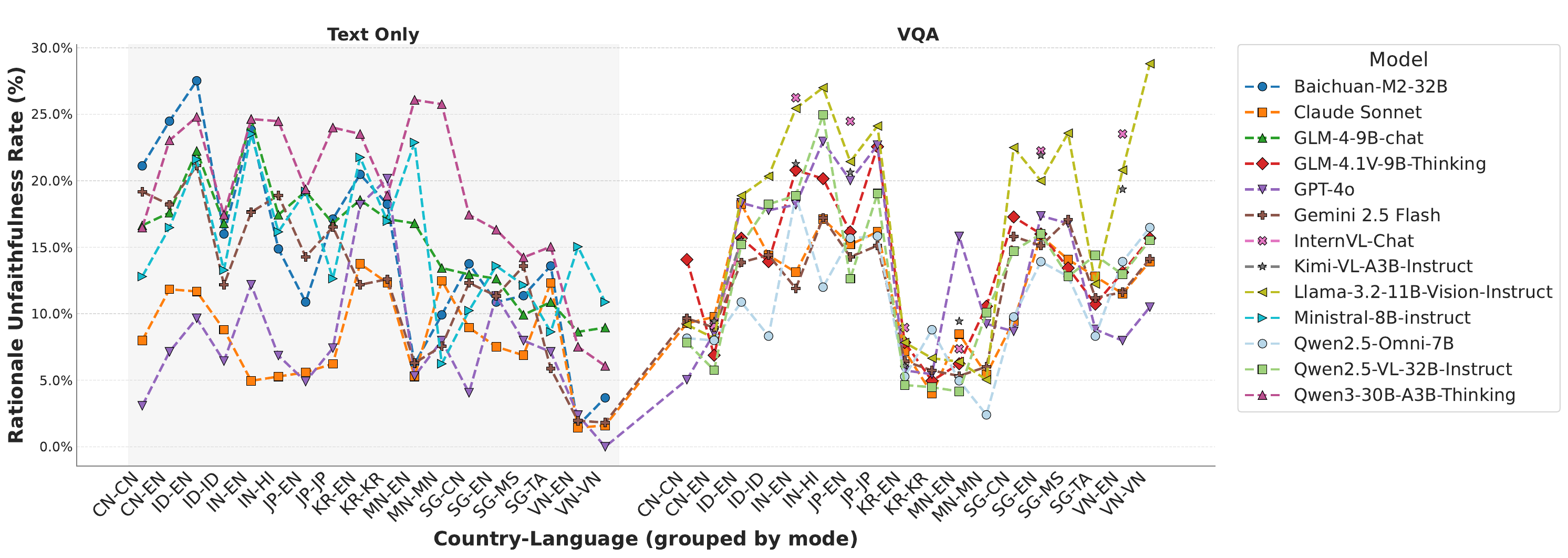}
    \label{fig:vqa and text-only}
  \end{subfigure}
  \captionsetup{skip=2pt}
  \caption{\textbf{Rationale Unfaithfulness Rates of LLMs across text-only and VQA.} Similar trends are observed for Rephrase VQA and Spoken QA; detailed results are provided in Appendix~\ref{sec:RUR}.}
  \label{fig:difference}
  \vspace{-0.7cm}
\end{figure}

\subsection{Cultural Awareness Grounding Validation}
\label{sec:Cultural Awareness Grounding Validation}
MCQs are convenient but can be solved via shortcuts (e.g., option elimination) rather than grounded knowledge~\citep{myung2025blendbenchmarkllmseveryday,romero2024cvqaculturallydiversemultilingualvisual,wang2024seaevalmultilingualfoundationmodels,wang2025generalizationvsmemorizationtracing,hartmann2023sokmemorizationgeneralpurposelarge}. We adopt two measures: (i) \textbf{Retained background knowledge:} Each item accompanied by itsemph{knowledge points} (supporting evidence). (ii) \textbf{Explainable responses:} During testing, models must provide a textual rationale for their choice.

We use a LLM-as-Judge approach to verify whether, \emph{given a correct answer}, the model’s explanation matches the item’s knowledge points. To reduce variability across judges, we require explanations in English. The human consistency checks, the multimodel consistency evaluation, and all LLM-as-Judge parameter settings are described in Appendix~\ref{sec:LLM-as-Judge}. Figure~\ref{fig:difference} reports the \emph{Rationale Unfaithfulness Rate} (RUR) for text-only and VQA items, defined as the proportion of correct answers whose explanations contradict or omit the required knowledge. Despite strong overall accuracy, proprietary models (Claude, GPT, and Gemini) still show RUR values between 5\% and 20\%. The issue is more pronounced for open-source models: Llama-3.2-11B-Vision-Instruct and Qwen3-30B-A3B-Thinking have the highest RUR, with Llama’s rate particularly elevated on non-English inputs. Qwen3-30B-A3B-Thinking often produces long explanations that contain hallucinations and sometimes derives the correct option from premises that contradict the ground truth, indicating reliance on generic heuristics rather than culturally grounded reasoning.

These results suggest that MCQs alone do not reliably measure cultural understanding (especially
for open-source models) and should be augmented with culture-grounded verification,
including knowledge-point checks and rationale assessment.

\subsection{Cultural-Awareness Consistency Across Languages }
\vspace{0pt}
\begin{figure}[t!]
  \centering
  \begin{subfigure}{0.43\linewidth}
    \centering
    \includegraphics[trim=18 5 18 8, clip,width=\linewidth]{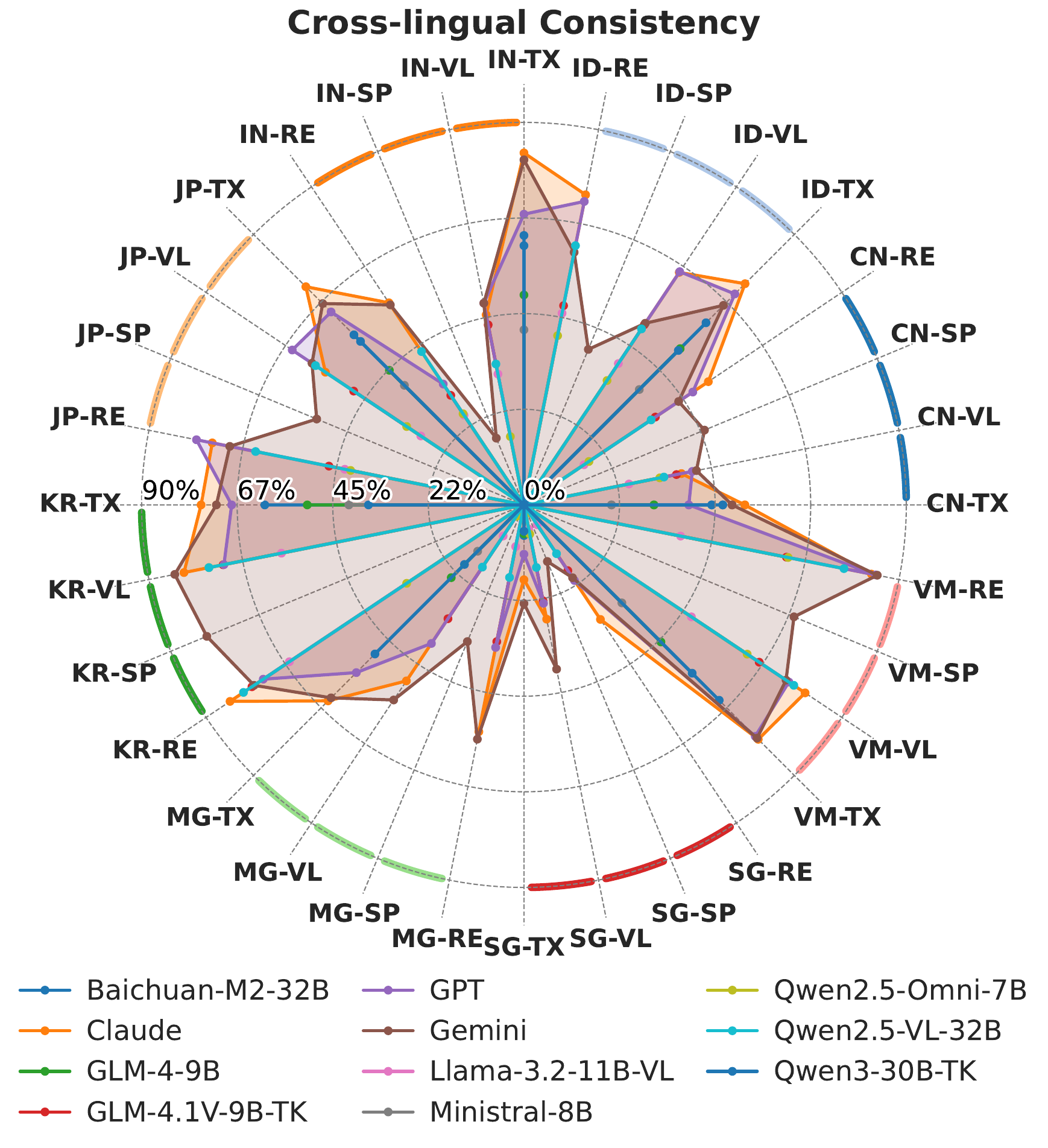}
    \captionsetup{skip=2pt}
    \caption{} 
    \label{fig:crosslingual}
  \end{subfigure}
  \begin{subfigure}{0.415\linewidth}
    \centering
    \includegraphics[trim=10 5 10 8, clip,width=\linewidth]{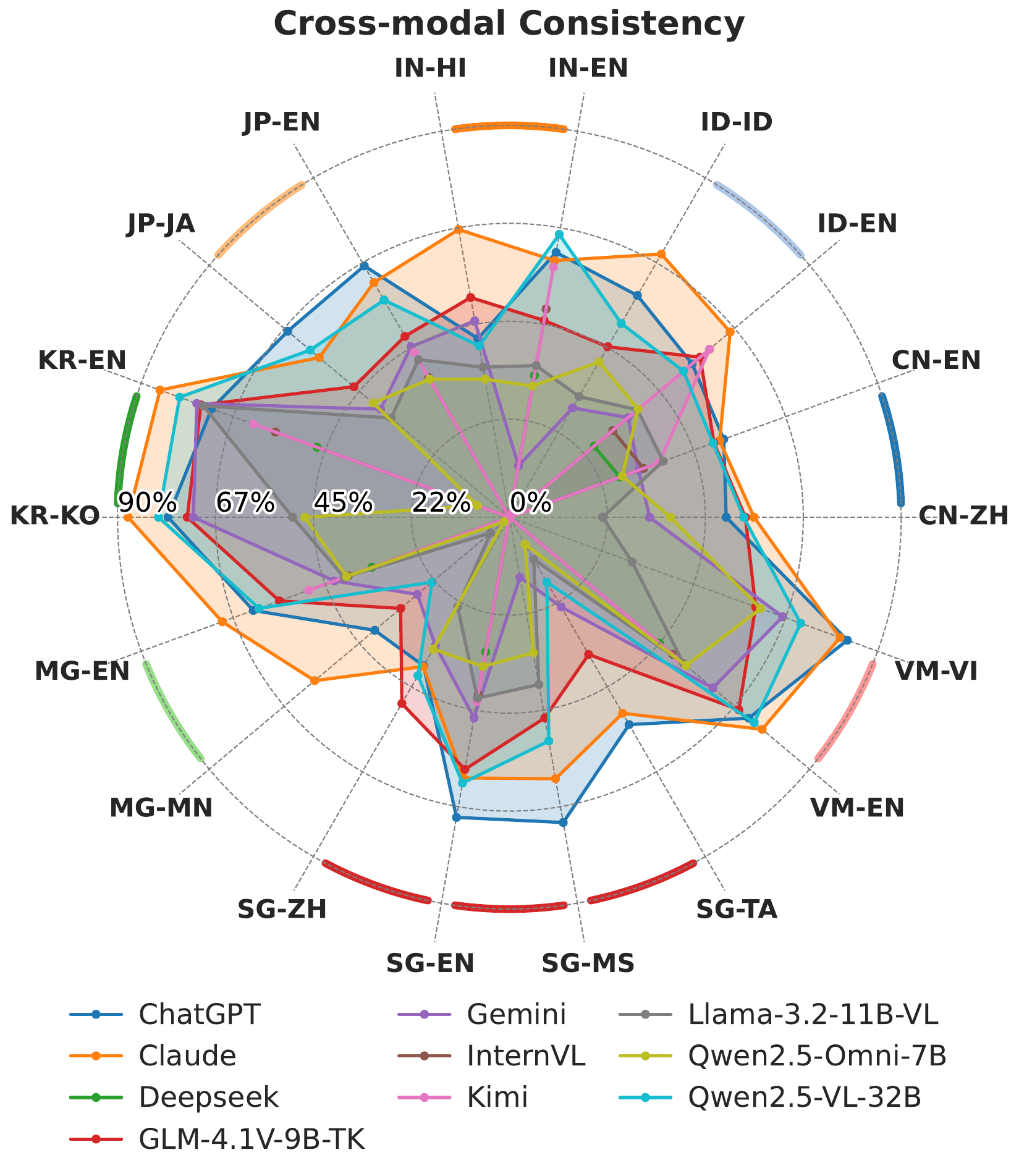}
    \captionsetup{skip=2pt}
    \caption{} 
    \label{fig:crossmodal}
  \end{subfigure}
  \captionsetup{skip=2pt}
  \caption{\textbf{(a) Cross-lingual consistency with fixed country and modality and (b) cross-modal consistency with fixed language and country.} TX/VL/RE/SP represent text QA, visual QA, rephrase QA, and speech QA.}
  \label{fig:CROSS}
  \vspace{-0.5cm}
\end{figure}

Cross-lingual cultural-awareness consistency is defined as the degree to which a model gives consistent outputs to semantically equivalent prompts posed in different languages, irrespective of answer correctness, which can be expressed as Eq. \ref{equation:consistency} \citep{wang2024seaevalmultilingualfoundationmodels}. 

{ 
\setlength{\abovedisplayskip}{-5pt} 
\setlength{\abovedisplayshortskip}{-5pt} 
\setlength{\belowdisplayskip}{-20pt}
\begin{equation}
\small
\mathrm{Consistency}_s =
\frac{1}{N}
\sum\nolimits_{i=1}^{N}
\frac{1}{\binom{m_i}{s}}
\sum_{\substack{S \subseteq L_i, |S|=s}}
\mathbf{1}\!\left( \bigl|\{ a_i^{\ell} : \ell \in S \}\bigr| = 1 \right).
\label{equation:consistency}
\end{equation}
}
where $L_i$ denotes the set of languages available for question $i$, and $m_i = |L_i|$ is its cardinality; $S \subseteq L_i$ with $|S| = s$ denotes any size-$s$ language subset; $a_i^{\ell}$ is the model’s answer to question $i$ when prompted in language $\ell$; $\mathbf{1}(\cdot)$ is the indicator function; and $\bigl|\{a_i^{\ell} : \ell \in S\}\bigr| = 1$ asserts that all answers within $S$ are identical. If $m_i < s$, there are no valid subsets and item $i$ contributes zero. 

As shown in Figure~\ref{fig:crosslingual}, contemporary multilingual LLMs exhibit weak cross-lingual consistency on culturally grounded tasks in both text-only and VQA settings. The weakness is most evident for language pairs with large resource gaps. For Mongolian culture, the disparity between Mongolian and English yields only 65.2\% consistency for Claude, while all open-source models remain below 50\%. By contrast, Korean culture shows higher consistency, plausibly reflecting the global diffusion of contemporary Korean media and the resulting multilingual exposure to related knowledge~\citep{jang2024more,dal2018analysis}. Consistency also declines sharply as the number of evaluated languages increases. For Singapore-related items, when Chinese, English, Tamil, and Malay are assessed jointly, the maximum consistency does not exceed 45\% (Gemini on VQA), despite relatively high pairwise values of 60.60\% (EN–TA), 64.20\% (EN–MS), and 55.20\% (EN–ZH). In some culturally challenging cases, visual context can partially bridge languages: for Indian culture, GLM-4.1 achieves 44.20\% cross-lingual consistency on Hindi VQA, which is 13.2 points higher than its rephrased text-only counterpart, although both remain low.

From these observations, we observe that the consistency between languages depends on the data resources and cultural exposure. Resource asymmetry degrades consistency, whereas cultural prominence helps. Consistency decays nonlinearly as more languages are considered, and strong pairwise agreement does not guarantee multi-language coherence. Visual cues can narrow gaps in certain settings, but are insufficient to overcome the structural limitations of low-resource languages.

\subsection{Cultural-Awareness Consistency Across Modalities}
Cross-modal cultural-awareness consistency evaluates whether a model gives the same output for semantically equivalent queries presented in different modalities. Figure~\ref{fig:crossmodal} shows that, across the eight Asian countries, the pattern largely matches the cross-lingual case: averaged over models, cross-lingual consistency is 48\%, and cross-modal consistency rarely exceeds 67\%. This gap indicates asymmetric transfer of cultural knowledge across modalities. Under low-resource language settings, almost all models struggle to maintain stable cross-modal answers. Within the same national context, medium- to high-resource local languages typically yield higher cross-modal consistency than English. To examine the observed ordering \emph{text-only} $>$ \emph{VQA} $>$ \emph{spoken QA} (see Section~\ref{sec:cultural Awareness Disparity}), we conduct a detailed error analysis with Qwen2.5-VL-32B-Instruct. Because spoken QA adds additional complexities (noise, accents, intonation), our analysis in this section focuses on VQA versus text-only performance; we leave a fuller study of speech to future work.

\begin{figure*}[t!]
  \centering
  \includegraphics[trim=0 350 0 0, clip, width=\textwidth]{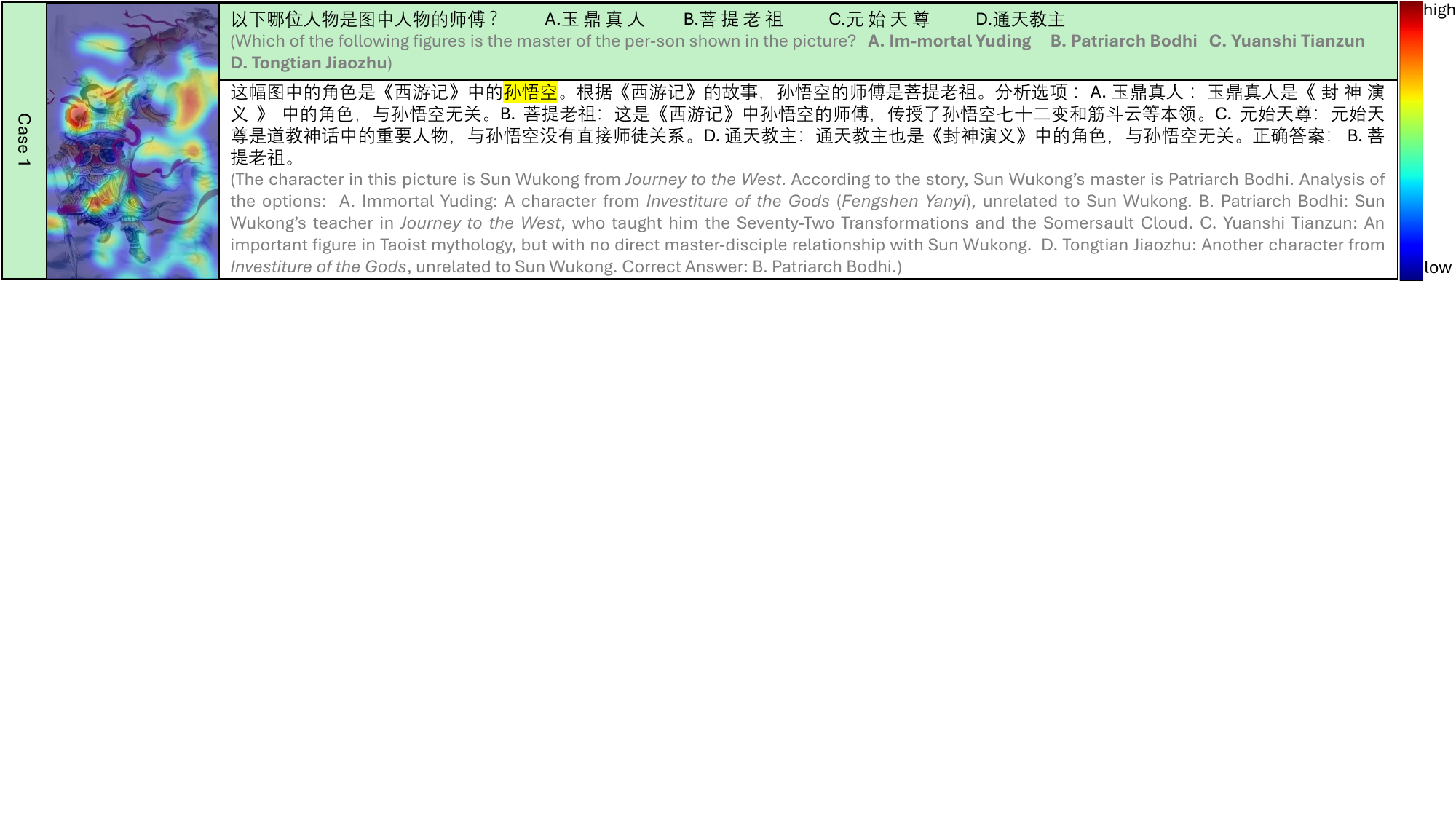}
  \captionsetup{skip=2pt}
  \caption{\textbf{Attention heatmap visualization over image regions during incorrect model answers.} Color scale from \BlueColorIcon blue (low) to \RedColorIcon red (high) indicates increasing model attention.}
  \label{fig:image hotmap}
  \vspace{-0.5cm}
\end{figure*}

We specifically isolate instances where the model succeeded with pure text input but failed in the VQA context. Apart from the most common errors arising from a lack of understanding of culture-related image contexts, our analysis reveals two additional predominant categories of errors:

\vspace{4pt}
\textbf{The pitfall of prompt-guided selective attention.} Models often tend to focus predominantly on explicitly mentioned objects in prompts, whereas cultural VQA requires a more nuanced ability to identify culture-specific visual cues within images. To validate whether the models' focuses are truly culture-specific, we extract visual evidence using answer-conditioned multi-layer Grad-CAM \citep{Selvaraju_2019}, interpolating and mapping the resulting heatmaps back to the original image for visualization. Specifically, given an image $v$, a textual prompt $x$ and an answer $y_{a:b}=(y_a,\dots,y_b)$ autoregressively produced by a model with parameters $\theta$, we define an answer-conditioned objective on the log-likelihood of the answer tokens as Eq. \ref{equation2}. Here we use the token-sum objective. For memory efficiency, only the visual tower is set to require gradients. We denote the forward activation of the $\ell$-th visual block as $A(\ell) \in \mathbb{R}^{C \times H \times W}$, the gradient of the block is $\mathbf{G}^{(\ell)}$. A per-layer Grad-CAM is built via channelwise inner product followed by $\operatorname{ReLU}$ (Eq.~\ref{eq:matrix_M_relu}).
\vspace{-5pt}

\noindent
\begin{minipage}{0.45\linewidth}
\small
\vspace*{-0.2\baselineskip}
\begin{equation}
    J_{\text{sum}}
    = -\sum\nolimits_{t=a}^{b} \log p_\theta\!\left(y_t \mid v, x, y_{<t}\right)
    \label{equation2}
\end{equation}
\end{minipage}\hfill
\begin{minipage}{0.5\linewidth}
\small
\vspace*{-0.2\baselineskip}
\begin{equation}\label{eq:matrix_M_relu}
\mathbf{M}^{(\ell)} = \operatorname{ReLU}\!\Big(\sum\nolimits_{c=1}^{C}
\mathbf{G}^{(\ell)}_c \odot \mathbf{A}^{(\ell)}_c\Big)\in \mathbb{R}^{H\times W}
\end{equation}
\end{minipage}

After lightweight smoothing and normalization of M($\ell$), the resulting $\hat{M}^{(\ell)}$ then undergoes cross-layer aggregation and bilinear interpolation upsampling on the $k$ blocks at the end of the visual tower, as defined by Eqs. \ref{eq:matrix_M_agg} and \ref{eq:matrix_M_tilde}, to the original image resolution. For more robust results, we specifically choose the last 3 blocks and do cross-layer aggregation by averaging them. 
\vspace{3pt}

\begin{minipage}[t]{0.32\linewidth}
\small
\vspace*{-0.8\baselineskip}
\begin{equation}\label{eq:matrix_M_agg}
\mathbf{M} = \operatorname{Agg}_{\ell\in\mathcal{L}}\big(\hat{\mathbf{M}}^{(\ell)}\big)
\end{equation}
\end{minipage}\hfill
\begin{minipage}[t]{0.64\linewidth}
\small
\vspace*{-0.8\baselineskip}
\begin{equation}\label{eq:matrix_M_tilde}
\tilde{\mathbf{M}} =
\operatorname{Bilinear}\!\left(\mathbf{M},\, H_{\text{img}},\, W_{\text{img}}\right)
\in[0,1]^{H_{\text{img}}\times W_{\text{img}}}.
\end{equation}
\end{minipage}
\vspace{5pt}

This phenomenon is clearly revealed by the heatmaps in Figure \ref{fig:image hotmap},  showing how the model focuses on subjects explicitly mentioned in the prompt when answering questions. However, this selective attention can unfortunately lead the model to overlook other critically important local details within the image, consequently resulting in erroneous inferences.
In Case 1 (Fig.~\ref{fig:image hotmap}), when the model is presented with the question ``Which of the following figures is the master of the person shown in the picture?'' in Chinese, its attention is predominantly drawn to the figure on the left side of the image, who is explicitly referenced in the prompt. However, the model overlooks the Howling Celestial Dog in the upper-right corner of the image, a crucial clue for identifying Erlang Shen. Additional attention-visualization examples and analyses are given in Appendix~\ref{sec: Attention heatmaps}. 

\begin{figure*}[t!]
  \centering
  \includegraphics[trim=0 320 0 0, clip,width=\textwidth]{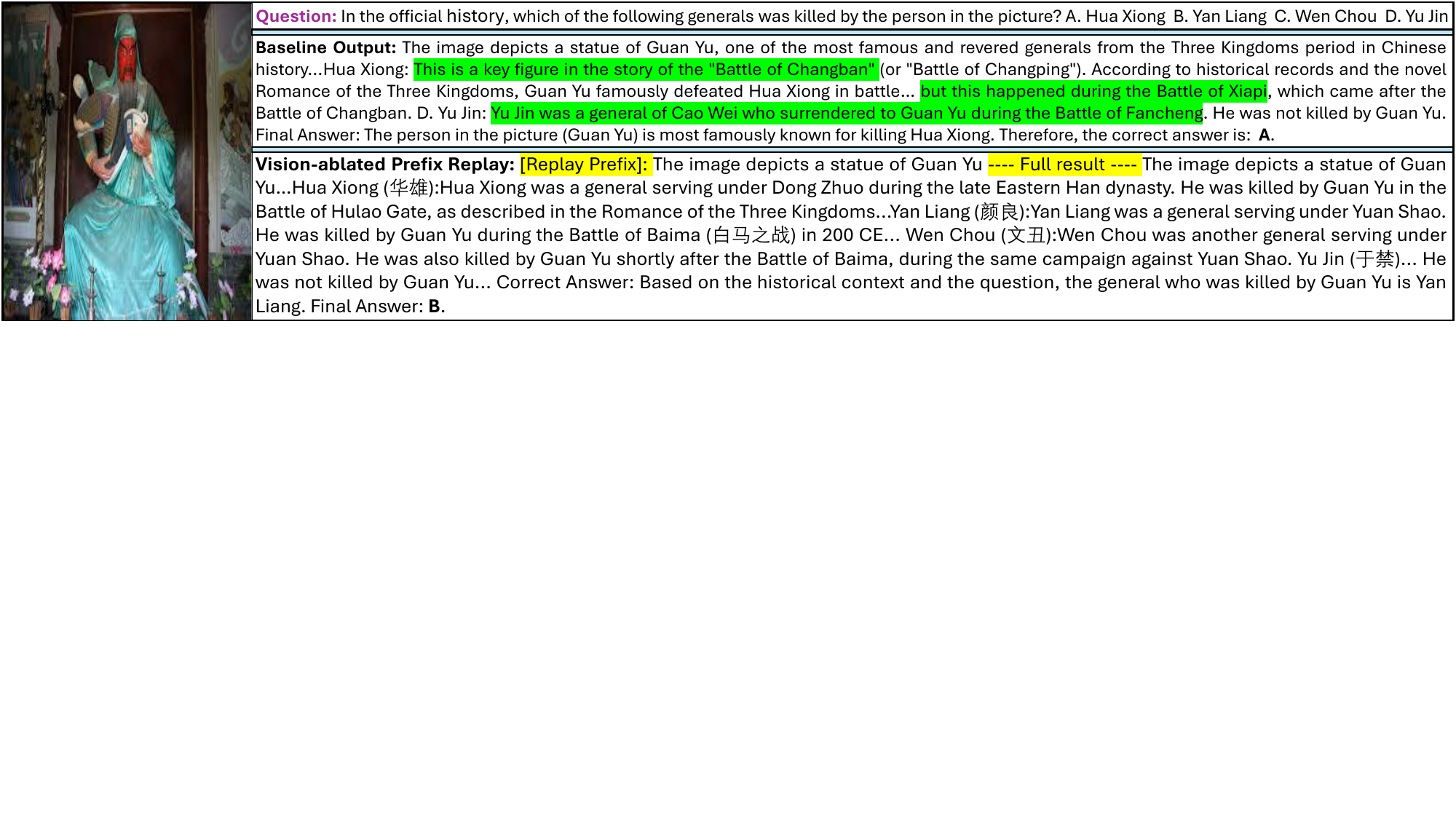}
  \captionsetup{skip=2pt}
  \caption{\textbf{Comparison of baseline vs. ``VPR" outputs, with hallucinations highlighted in green.}}
  \label{fig:replay result}
  \vspace{-0.5cm}
\end{figure*}

\vspace{4pt}
\textbf{Image tokens contribute to reasoning hallucinations.} We find that visual content increases reasoning hallucinations in models compared to text-only QA, despite accurate image recognition capabilities. In Figure \ref{fig:replay result}, while the baseline model correctly identifies ``Guan Yu'', it still produces multiple reasoning hallucinations (highlighted in green). However, text-only Rephrase VQA queries show no such hallucinations (Appendix~\ref{sec: Vision-ablated Prefix Replay}), indicating that reasoning errors likely originate from image tokens and suggest modality-specific bias in multimodal reasoning. To test this hypothesis, we propose Vision-ablated Prefix Replay (VPR), which generates image descriptions then removes visual conditions while maintaining fixed prefixes for subsequent reasoning (details in Appendix \ref{sec: Vision-ablated Prefix Replay}). In Figure \ref{fig:replay result}, VPR conditions the model's reasoning on the generated caption `The image depicts a statue of Guan Yu' while discarding visual tokens, eliminating hallucinations and producing the correct answer. Furthermore, we select 50 questions where VQA answers contained hallucinations but corresponding text-only queries were correct. VPR eliminates hallucinations and produced correct answers for 19 of these cases (38\%), supporting our hypothesis.

Multimodal models demonstrate significant inconsistency in cultural awareness across modalities, indicating flawed cultural knowledge transfer. In VQA, this deficiency stems from two core issues: ``\textit{selective attention pitfall}'' where models over-focus on text-prompted subjects while missing key visual cues, and visual token-induced ``\textit{reasoning hallucinations}''.

\subsection{Cultural Knowledge Generalization}
\label{sec:Cultural Knowledge Generalization}

Prior work \citep{balepur-etal-2024-artifacts,molfese2025right,zheng2023large} suggests that scaling increases LLMs’ factual memory but not genuine logical generalization. To distinguish whether cultural multi-step reasoning errors stem from knowledge gaps or generalization deficits, we conduct a deconstruction study. We decompose each question into atomic sub-questions testing single knowledge points, which models answer first (see Table \ref{table:multi-steps} for an example). We then evaluate the original question under two conditions: (i) with in-context ``sub-question → model answer'' pairs, and (ii) from scratch. If models solve all subquestions but fail the original question from scratch, they possess the knowledge but cannot transfer it, indicating generalization deficits. If they err on sub-questions, failure likely reflects missing culture-specific knowledge.

We investigate the failure modes of Claude Sonnet 4 (closed-source) and Qwen3-30B-A3B-Thinking (open-source) on multi-step reasoning tasks in English. Our experiment uses problems that each model initially failed (232 for Claude, 271 for Qwen), decomposing each into 3-8 sub-questions with ground-truth answers. An LLM (GPT-4o) is used to judge the correctness of the model’s answer to each step. The results for Claude and Qwen are as follows:
\begin{itemize}[leftmargin=*,topsep=2pt]
\setlength{\itemsep}{0pt}
\setlength{\parskip}{0pt}
\setlength{\parsep}{0pt}
  \item \textbf{Successful Correction:} 35.3\% (Claude), 18.5\% (Qwen)  of problems were resolved, with all sub-questions and the final contextualized answer being correct.
  \item \textbf{Integration Failure:} 2.6\% (Claude), 10.7\%(Qwen) of cases failed at the final synthesis step, despite all sub-questions being answered correctly.
  \item \textbf{Sub-question Failure:} 62.1\% (Claude), 70.8\% (Qwen) cases failed at the initial sub-question reasoning stage.
\end{itemize}
\noindent
Obviously, the transfer and generalization of cultural knowledge pose a significant challenge for language models, irrespective of their scale. Claude corrects 35.3\% of initial errors through stepwise decomposition-reintegration evaluation, showing broad knowledge coverage but limited cultural generalization. Qwen's performance is constrained by knowledge gaps and poor transferability. While 29.2\% of multi-step problems have all sub-questions answered correctly, 10.7\% fail at reintegration, indicating weaker cultural generalization capabilities.

\section{Conclusion}
We introduced MMA-ASIA, a tri-modal (text, image, speech), multilingual benchmark and framework for evaluating cultural awareness in LLMs across 8 Asian countries and 10 languages. Our contributions include an aligned, human-curated dataset with substantial multi-step reasoning, a five-dimensional protocol that measures accuracy, cross-lingual and cross-modal consistency, cultural knowledge generalization, and grounding validity, and analysis tools that reveal shortcut use. Results show persistent data-driven cultural bias, uneven cross-lingual transfer, and fragile multimodal reasoning (selective visual attention and image-induced hallucinations). At the same time, accented speech can act as a useful cultural cue. Models also struggle to integrate known facts into multi-step reasoning, indicating a generalization bottleneck. We argue for consistency- and grounding-aware evaluation, as well as methods that strengthen cross-modal alignment and broaden high-quality coverage in low-resource languages. MMA-ASIA provides data, protocols, and baselines to track progress toward culturally reliable multimodal LLMs.

\bibliography{iclr2026_conference}
\bibliographystyle{iclr2026_conference}

\appendix
\section{Appendix}

\subsection{Annotator Demographic}
\label{sec:Annotator Demographic}

Our annotation team comprises members from eight different countries. All team members are native speakers of their local languages and proficient in English, with professional backgrounds in natural language processing or speech processing. Each annotator has lived in the respective country for more than ten years. Table \ref{tab:annotator_demographics} presents the annotators’ details and professional backgrounds; to protect privacy, we replace personal names with numeric identifiers within each country. After each team completes the first round of annotations, in-country linguistic experts conduct a data review. Table \ref{tab:reviewer_demographics} lists the language experts’ information; likewise, we anonymize personal names.

\begin{table}[htbp]
  \centering
   \caption{Data Annotator Demographics and Skills}
  \label{tab:annotator_demographics}
  \begin{tabular}{@{} l c l c l l l l @{}}
    \toprule
    \textbf{Country} & \textbf{ID} & \textbf{Gender} & \textbf{Age} & \textbf{Education} & \textbf{\begin{tabular}[c]{@{}l@{}}English \\ Prof.\end{tabular}} & \textbf{\begin{tabular}[c]{@{}l@{}}Local Lang. \\ Prof.\end{tabular}} & \textbf{\begin{tabular}[c]{@{}l@{}}Professional \\ Background\end{tabular}} \\
    \midrule
    \multirow{5}{*}{China} 
    & 1 & Male & 28 & Ph.D. & Proficient & Proficient & NLP \\
    & 2 & Male & 25 & Master Deg. & Proficient & Proficient & NLP \\
    & 3 & Male & 30 & Ph.D. & Proficient & Proficient & NLP \\
    & 4 & Male & 25 & Master Deg. & Familiar & Fluent & NLP \\
    & 5 & Male & 24 & Master Deg. & Familiar & Proficient & NLP \\
    \midrule
    \multirow{2}{*}{Japan} 
    & 1 & Male & 23 & Master Deg. & Proficient & Proficient & SP \\
    & 2 & Male & 35 & Ph.D. & Fluent & Proficient & SP \\
    \midrule
    \multirow{2}{*}{Mongolia} 
    & 1 & Male & 26 & Ph.D. & Proficient & Proficient & SP \\
    & 2 & Male & 25 & Master Deg. & Proficient & Proficient & SP \\
    \midrule
    \multirow{3}{*}{Korea} 
    & 1 & Male & 31 & Ph.D. & Proficient & Proficient & NLP \\
    & 2 & Male & 26 & Master Deg. & Proficient & Proficient & NLP \\
    & 3 & Female & 25 & Master Deg. & Proficient & Proficient & NLP \\
    \midrule
    \multirow{2}{*}{India} 
    & 1 & Male & 25 & Bachelor's Deg. & Proficient & Proficient & NLP \\
    & 2 & Male & 31 & Ph.D. & Proficient & Proficient & NLP \\
    \midrule
    \multirow{5}{*}{Vietnam} 
    & 1 & Male & 21 & Bachelor's Deg. & Proficient & Proficient & SP \\
    & 2 & Male & 20 & Bachelor's Deg. & Proficient & Proficient & SP \\
    & 3 & Male & 21 & Bachelor's Deg. & Proficient & Proficient & SP \\
    & 4 & Male & 21 & Bachelor's Deg. & Proficient & Proficient & SP \\
    & 5 & Male & 29 & Master Deg. & Proficient & Proficient & SP \\
    \midrule
    \multirow{2}{*}{Indonesia} 
    & 1 & Female & 21 & Bachelor's Deg. & Proficient & Proficient & NLP \\
    & 2 & Female & 26 & Ph.D. & Proficient & Proficient & NLP \\
    \midrule
    \multirow{5}{*}{Singapore} 
    & 1 & Female & 26 & Ph.D. & Proficient & Proficient & NLP \\
    & 2 & Female & 18 & Bachelor's Deg. & Proficient & Proficient & NLP \\
    & 3 & Male & 18 & Bachelor's Deg. & Proficient & Proficient & NLP \\
    & 4 & Male & 21 & Bachelor's Deg. & Proficient & Proficient & NLP \\
    & 5 & Female & 28 & Ph.D. & Proficient & Proficient & SP \\
    \bottomrule
  \end{tabular}
  \par
  \small\textit{Note: NLP: Natural Language Processing; SP: Speech Processing.}
\end{table}

\begin{table}[htbp]
  \centering
  \caption{Data Reviewer Demographics and Skills}
  \label{tab:reviewer_demographics}
  \begin{tabular}{@{} l c l c l l l @{}}
    \toprule
    \textbf{Country} & \textbf{ID} & \textbf{Gender} & \textbf{Age} & \textbf{Education} & \textbf{\begin{tabular}[c]{@{}l@{}}English \\ Prof.\end{tabular}} & \textbf{\begin{tabular}[c]{@{}l@{}}Local Lang. \\ Prof.\end{tabular}} \\
    \midrule
    China     & 1 & Female & 30 & Master Deg.     & Proficient & Proficient \\
    Japan     & 1 & Female & 32 & Master Deg.     & Proficient & Proficient \\
    Mongolia  & 1 & Male   & 25 & Master Deg.     & Proficient & Proficient \\
    Korea     & 1 & Male   & 54 & Bachelor's Deg. & Proficient & Proficient \\
    India     & 1 & Male   & 28 & Master Deg.     & Proficient & Proficient \\
    Vietnam   & 1 & Male   & 27 & Master Deg.     & Proficient & Proficient \\
    Indonesia & 1 & Female & 28 & Bachelor's Deg. & Proficient & Proficient \\
    \midrule
    \multirow{4}{*}{Singapore} 
    & 1 & Female & 26 & Bachelor's Deg. & Proficient & Proficient \\
    & 2 & Female & 22 & Bachelor's Deg. & Proficient & Proficient \\
    & 3 & Male   & 30 & Ph.D.           & Proficient & Proficient \\
    & 4 & Male   & 26 & Ph.D.           & Proficient & Proficient \\
    \bottomrule
  \end{tabular}
\end{table}

\subsection{Data Statistics}
\label{sec:data statistic}

\begin{figure*}[t]
  \centering
  \includegraphics[trim=0 0 0 0, clip,width=\textwidth]{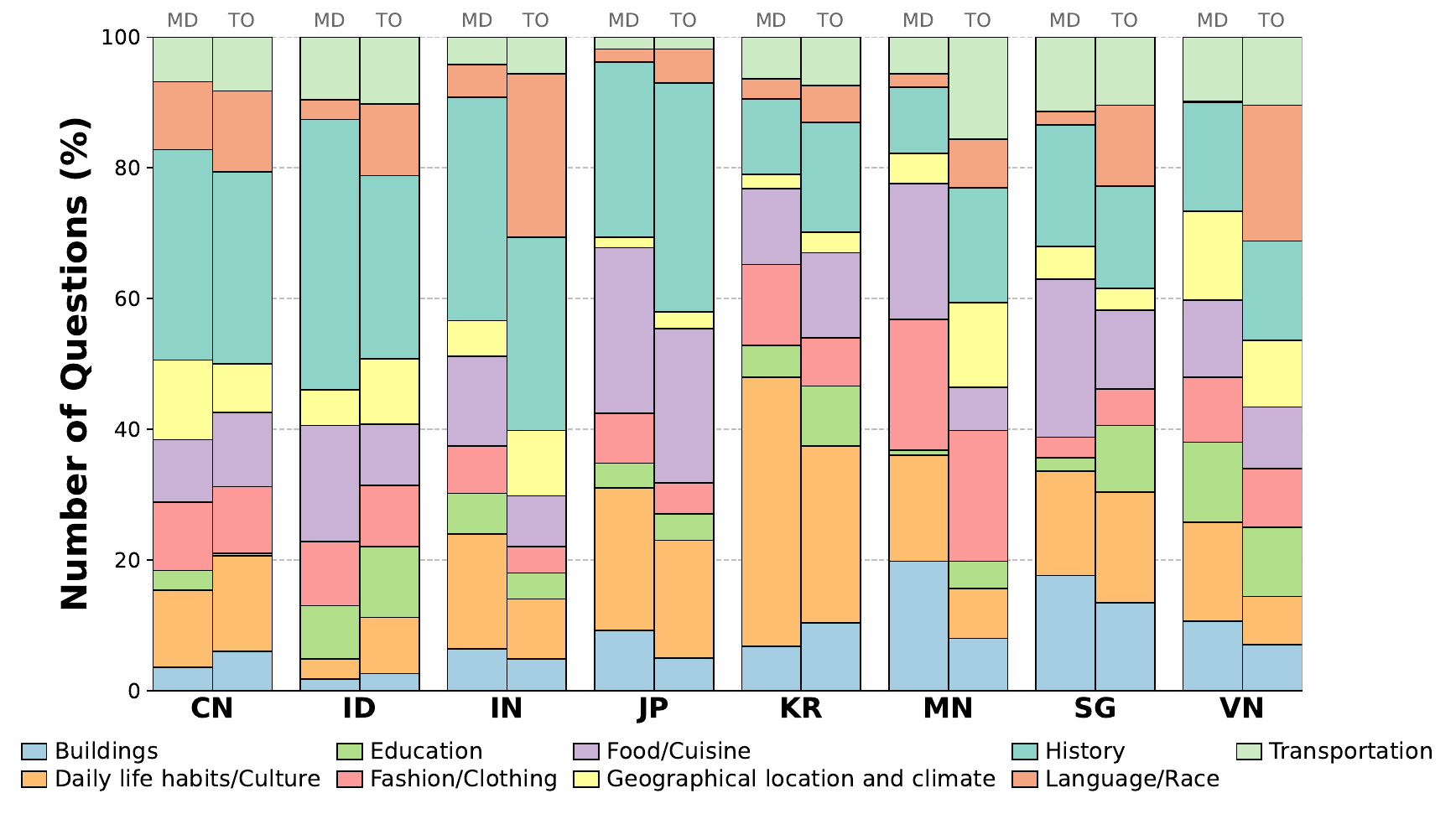}
  \caption{Distribution of question categories across countries and modalities}
  \label{fig:category}
\end{figure*}

\begin{figure*}[t]
  \centering
  \includegraphics[trim=0 0 0 0, clip,width=\textwidth]{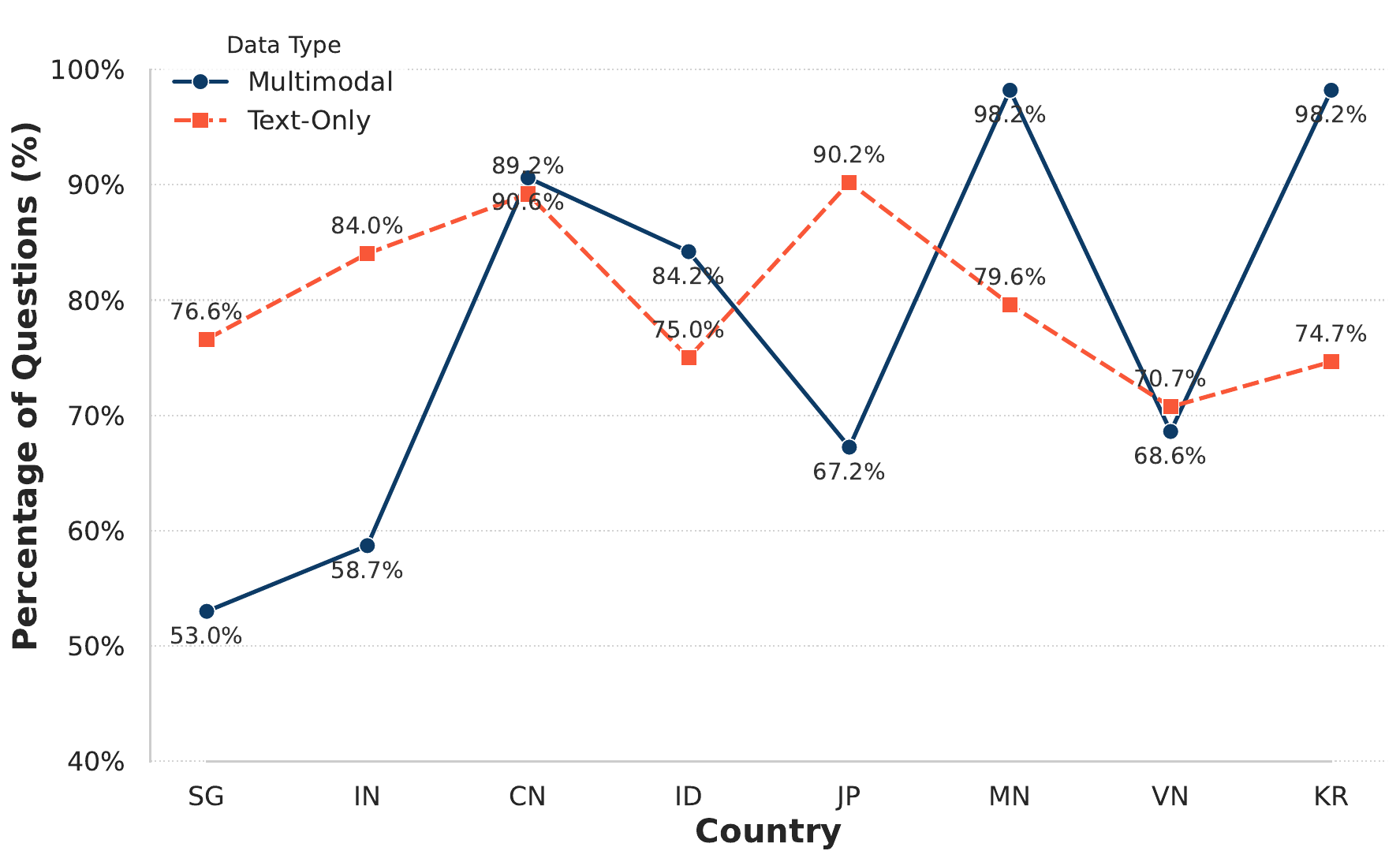}
  \caption{Multi-step reasoning question proportions across countries}
  \label{fig:multi-setps questions}
\end{figure*}

Under the MMA-ASIA framework, the dataset covers 8 countries and 10 languages, with each country’s split presented in both English and its local language, totaling 27,000 questions. Table \ref{tab:local_languages} lists the countries included in our dataset and their corresponding local language(s). Over 79\% of all items are multi-step cultural reasoning questions. We define a multi-step cultural reasoning item as one whose solution requires sequential derivation and/or synthesis from at least two independent knowledge components, rather than mere recall or paraphrase of a single cultural fact. Table~\ref{Reasoning case} presents an example multi-step question with its analysis. The proportion of multi-step items by country and modality is shown in Figure~\ref{fig:multi-setps questions}. The dataset spans nine categories—\emph{Daily Life/Culture, Food/Cuisine, Transportation, Buildings, History, Geographical Location \& Climate, Education, Fashion/Clothing, and Language/Ethnicity}—with per-country category distributions summarized in Figure~\ref{fig:category}.

\begin{table}[htbp]
  \centering
  \caption{Local Languages by Country}
  \label{tab:local_languages}
  \begin{tabular}{@{} l l @{}}
    \toprule
    \textbf{Country} & \textbf{Local Language} \\
    \midrule
    China     & Chinese \\
    Singapore & \begin{tabular}[t]{@{}l@{}}English, Chinese, \\ Malay, Tamil\end{tabular} \\
    Japan     & Japanese \\
    Korea     & Korean \\
    India     & Hindi \\
    Indonesia & Indonesian \\
    Vietnam   & Vietnamese \\
    Mongolia  & Mongolian \\
    \bottomrule
  \end{tabular}
\end{table}

The following outlines the potential contents included in each category.

\begin{itemize}[leftmargin=*, itemsep=2pt]
  \item \textbf{Daily Life/Culture}:
  Covers everyday etiquette, customs, and values—greeting practices, daily routines and social decorum, family and community interactions, as well as festivals and folk practices. It may also include consumption and leisure preferences, common life scenarios, and behavioral norms.

  \item \textbf{Food/Cuisine}:
  Regional cuisines, representative dishes and ingredients, cooking techniques, dietary taboos and table manners, utensils, and dining settings. Also includes festive foods, street snacks, and regional taste differences.

  \item \textbf{Transportation}:
  Transportation across historical periods and the evolution of vehicles, along with regional differences in modes of transport. May also cover landmark transit systems and commuting culture.

  \item \textbf{Buildings}:
  Traditional and modern architectural styles; religious and public buildings; housing forms and materials; city skylines and the preservation of historic districts. Can also address symbolic meanings in architecture and region-specific structural features.

  \item \textbf{History}:
  Major historical periods and events, notable figures and heritage sites, and how historical memory shapes contemporary society and culture. May also include colonial/independence histories and cultural change driven by migration and war.

  \item \textbf{Geographical Location \& Climate}:
  Landforms and terrain, climate zones and seasonal variation, natural resources, and ecosystems. Extends to lifestyle, clothing, and dietary adaptations shaped by geography and climate.

  \item \textbf{Education}:
  Renowned national works of literature, art, or music, as well as the structure of the education system and pathways to advancement, including stories surrounding prestigious institutions.

  \item \textbf{Fashion/Clothing}:
  Traditional attire and its ceremonial contexts; modern dress styles and aesthetic trends; accessories and color preferences; occupational/school uniforms and seasonal clothing. May also discuss cultural symbolism embedded in garments.

  \item \textbf{Language/Ethnicity}:
  Code-switching (where applicable); official and commonly used languages; dialects and accent features; writing systems and naming conventions; multi-ethnic compositions and cultural practices. Also includes politeness strategies in language and norms of cross-group communication.
\end{itemize}

\begin{table}[htbp]
\centering
\small
\caption{An example of multi-step reasoning question}
\label{Reasoning case}
\setlength{\tabcolsep}{4pt}
\begin{tabularx}{\textwidth}{@{}p{3.0cm} X@{}}
\toprule
\textbf{Question} &
An ancient tower became famous due to a poem by the Tang Dynasty poet Cui Hao. 
In which period was it proposed to be constructed with iron materials?
\textit{(A) 16th year of Guangxu, Qing Dynasty \quad
(B) 1st year of Jiaqing, Qing Dynasty \quad
(C) 7th year of Tongzhi, Qing Dynasty \quad
(D) 8th year of the Republic of China} \\
\midrule
\textbf{Reasoning Decomposition} &
\textbf{Step 1: Identify the tower.}
The poem is Cui Hao’s “Yellow Crane Tower,” so the tower is \emph{Yellow Crane Tower}.\par
\textbf{Step 2: Recall historical events.}
In the 10th year of Guangxu (1884) the tower was destroyed by fire.
In the 16th year of Guangxu (1890), Zhang Zhidong (Governor-General of Hubei and Hunan) first proposed rebuilding the tower using iron materials.\par
\textbf{Step 3: Match with the options.}
A (1890): \textbf{Correct}, matches the historical fact.
B (1796): Incorrect, too early.
C (1868): Incorrect, before the fire.
D (1919): Incorrect, after the Qing Dynasty and not the first proposal.\par
\textbf{Final Answer:} \textbf{A. 16th year of Guangxu, Qing Dynasty}.
\\
\bottomrule
\end{tabularx}
\end{table}

\begin{figure}[h]
  \centering

  \begin{subfigure}{\textwidth}
    \centering
    \includegraphics[trim=0 110 0 10, width=\linewidth]{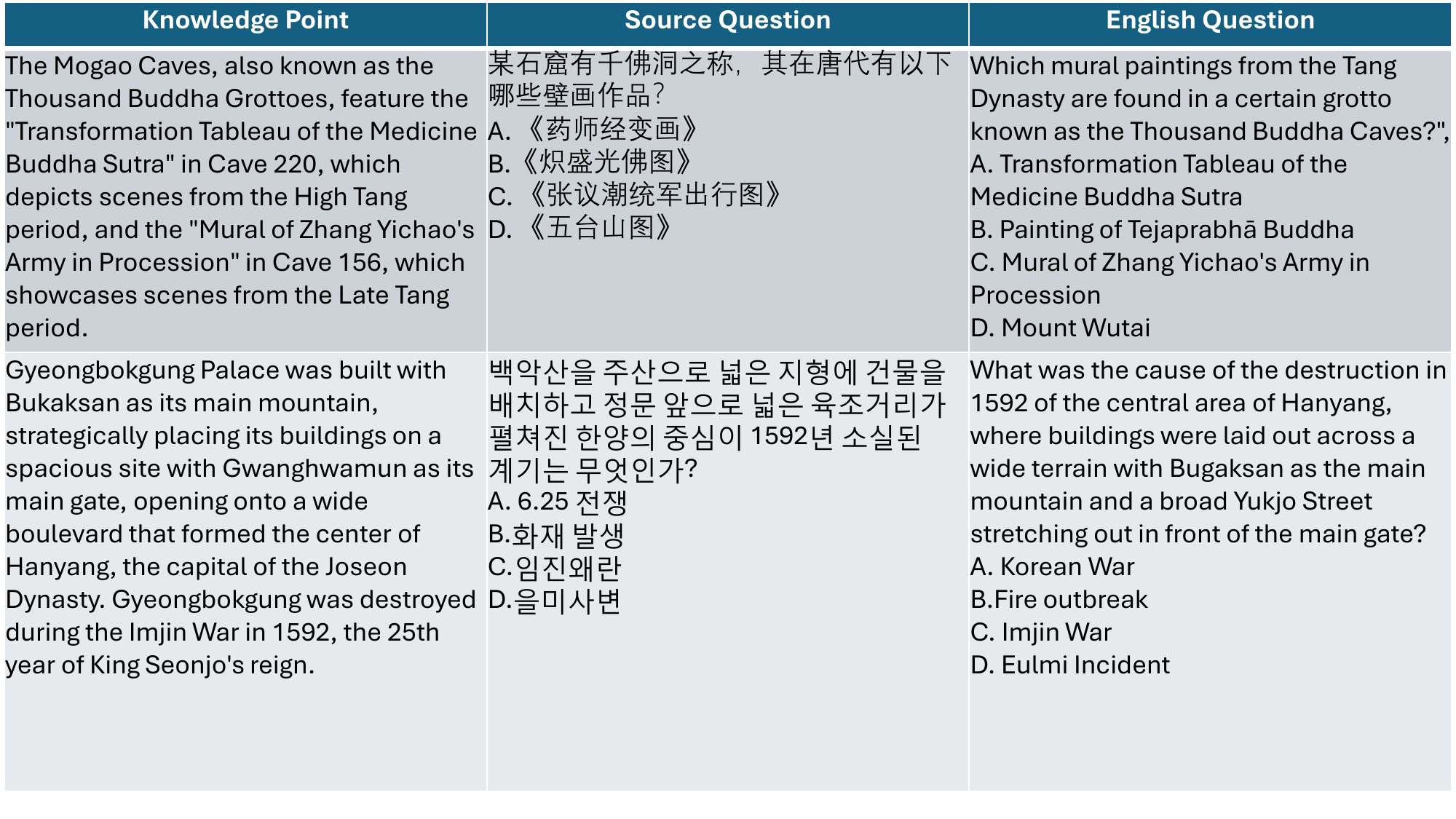}
    \label{fig:3}
  \end{subfigure}


  \begin{subfigure}{\textwidth}
    \centering
    \includegraphics[trim=0 64 0 10,width=\linewidth]{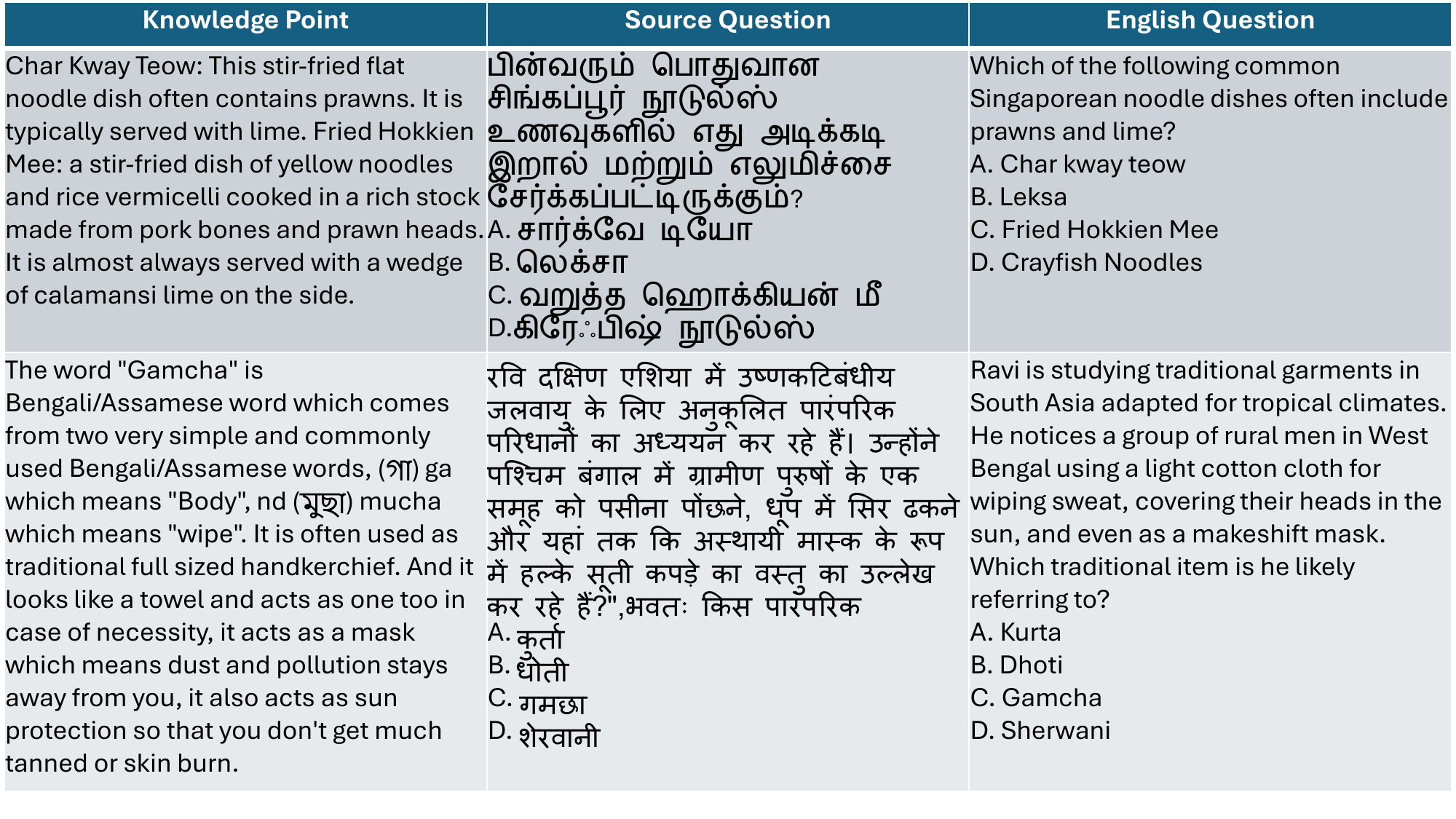}
    \label{fig:2}
  \end{subfigure}

  \caption{Text-Only Question Examples}
  \label{fig:1}
\end{figure}

\begin{figure*}[t]
  \centering
  \includegraphics[width=\textwidth]{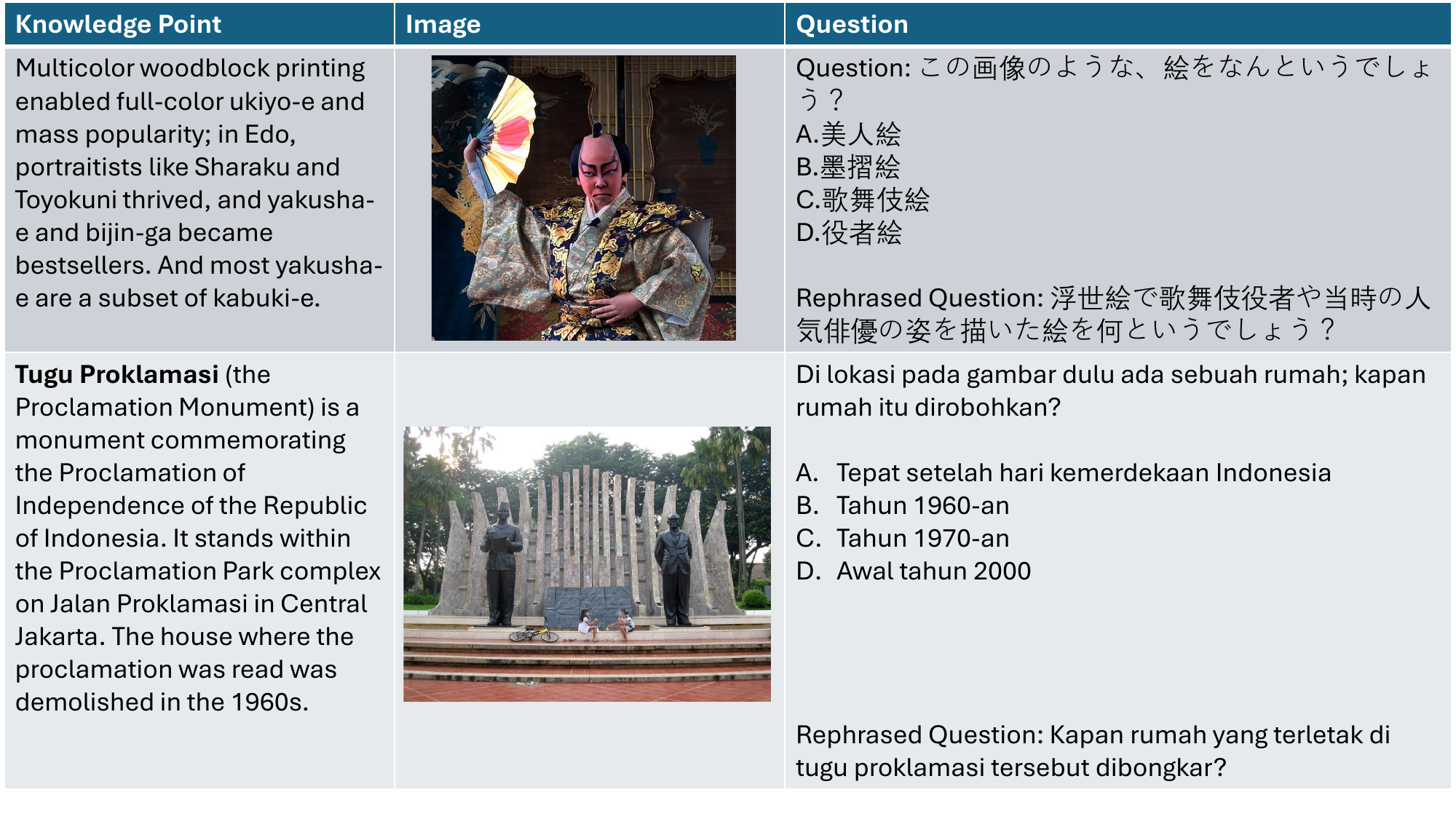}
  \caption{VQA/Rephrase VQA Question Examples}
  \label{fig:VQA_samples}
\end{figure*}

Figure~\ref{fig:1} and Figure~\ref{fig:VQA_samples} show some samples of our dataset.

\subsection{Annotation Guideline}
\label{sec:annotation guideline}

\begin{figure*}[h]
  \centering
  \includegraphics[angle=-90,origin=c,height=\textwidth, trim=40 20 40 20,clip]{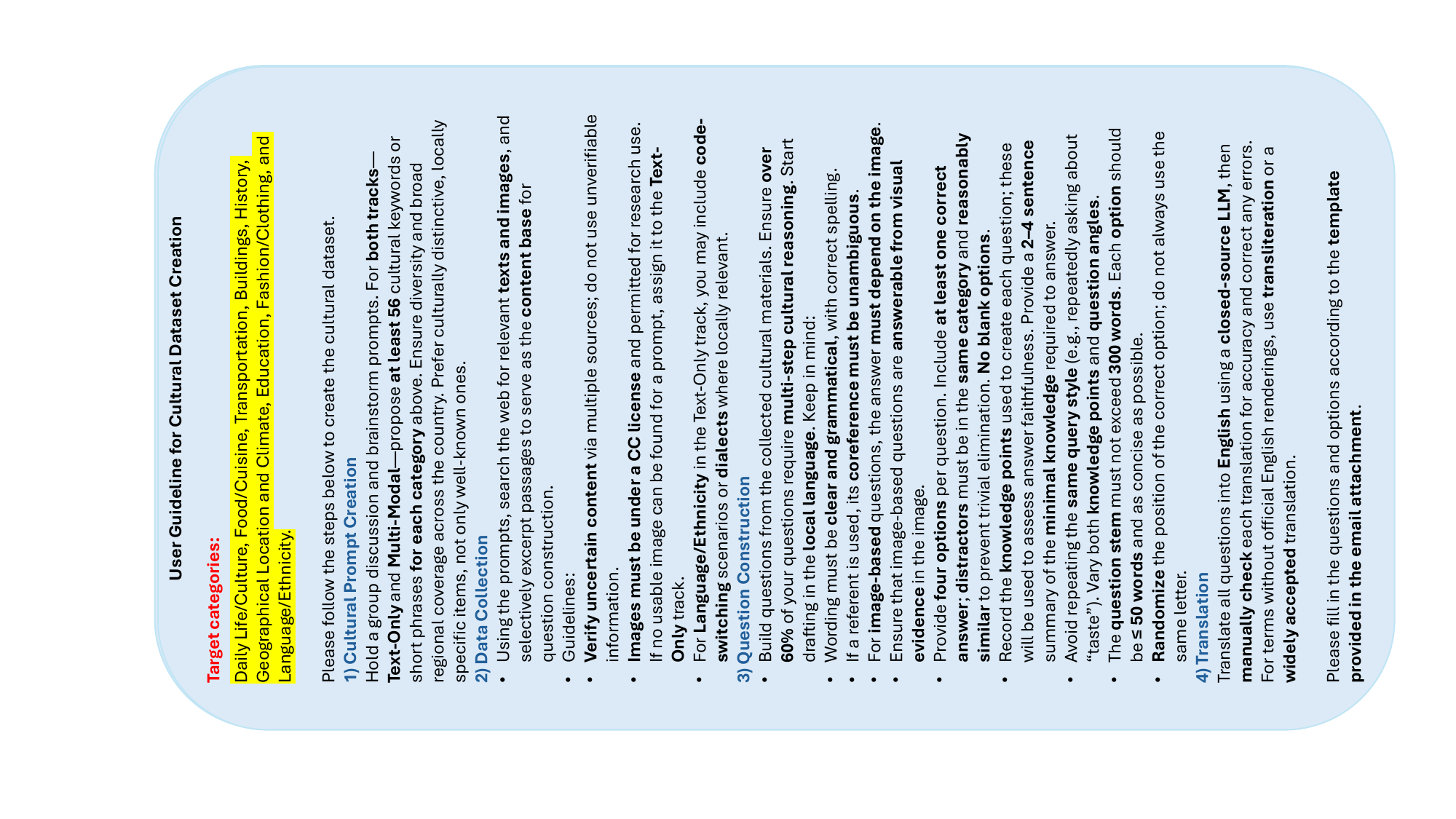}
  \caption{Annotator Guideline}
  \label{fig:annotator-guideline}
\end{figure*}

The content of the guideline distributed to annotators is shown in Figure~\ref{fig:annotator-guideline}. To minimize heuristic cues arising from non-cultural knowledge, we add a consistency constraint on distractors: they must belong to the same category as the correct option and closely resemble it in observable attributes and semantic representation. We also encourage each team to uncover cultural elements unique to their own country, rather than focusing only on widely known aspects. For the Language category, if code-switching is prevalent in the annotators’ country, we strongly encourage including such language-assessment examples in the Text-Only portion of the dataset. All content involving racism or hate speech is prohibited from inclusion in our dataset.

\subsection{Details of the TTS Tool and Procedure for Building the Speech Data}
\label{sec:TTS tools details}

Before generating the English audio, we standardized the input text through normalization of numbers and symbols, handling of abbreviations and special terms, and sentence segmentation.

To ensure high-quality speech synthesis, we employed CosyVoice\citep{du2025cosyvoice} for English audio generation. This tool supports voice cloning from sampled speakers, producing speech that preserves the timbre and accent of the reference voice. We collected representative recordings from native speakers across eight countries to capture diverse accents for speech synthesis. For standard English, we adopted CosyVoice’s built-in default English voice (English female voice). For non-English languages, CosyVoice was also used to generate Chinese, Japanese, and Korean audio. In addition, we employed in-house high-quality TTS systems built by different speech processing teams for Vietnamese, Tamil, Mongolian, and Malay, while the Coqui-ai TTS toolkit~\citep{coqui_tts_github} was used for Indonesian and Hindi.

Each generated audio sample was individually verified. When errors occurred—such as inappropriate pauses, missing segments, or mispronunciations—we first adjusted the input text and re-synthesized the audio, as TTS systems are often highly sensitive to textual variations. If repeated corrections still failed, we resorted to manual re-recording. Unlike in other language tasks, our requirement here was not fluency or naturalness, but rather clear articulation of the questions and answer options.

All audio files were standardized in WAV format with a 16 kHz sampling rate. And our speech generation uses two input types.
\begin{enumerate}
  \item Only the question stem from the \textit{Rephrase VQA (text-only)} item;
  \item The entire \textit{Rephrase VQA (text-only)} item, including the question and its answer options.
\end{enumerate}
Table~\ref{tab:TTS models} summarizes the models used for speech generation across different languages.

\begin{table}[h]
  \centering
    \caption{Models used for speech data generation}
  \label{tab:TTS models}
  \begin{tabularx}{\linewidth}{@{}l X@{}} 
    \toprule
    \textbf{Language} & \textbf{Model} \\
    \midrule
    English with accent, Korean, Japanese, Chinese & CosyVoice2-0.5B~\citep{du2025cosyvoice} \\
    English without accent & CosyVoice-300M~\citep{du2024cosyvoice} \\
    Vietnamese, Tamil, Mongolian, Malay & Internally developed TTS systems \\
    Indonesian & Indonesian TTS~\citep{https://doi.org/10.48550/arxiv.2106.06103} \\
    Hindi & AI4Bharat Indic-TTS~\citep{sankar2024indicvoices} \\
    \bottomrule
  \end{tabularx}
\end{table}

\subsection{Model Selection}
\label{sec:Model Selection}

\paragraph{Open-source multilingual text-only LLMs.} 
We evaluate Qwen3-30B-A3B-Thinking-2507~\cite{qwen3technicalreport}, Baichuan-M2-32B~\cite{baichuan-m2}, GLM-4-9B-Chat\cite{glm2024chatglm}, and Ministral-8B-Instruct~\cite{jiang2023mistral7b} for their multilingual capabilities.

\paragraph{Open-source multimodal LLMs.} We evaluate nine vision–language (and omni) models on image and text: Qwen2.5-VL-32B-Instruct~\citep{bai2025qwen25vltechnicalreport}, Llama-3.2-11B-Vision-Instruct~\citep{touvron2023llamaopenefficientfoundation}, Kimi-VL-A3B-Instruct~\citep{kimiteam2025kimivltechnicalreport}, DeepSeek-VL-Small~\citep{wu2024deepseekvl2mixtureofexpertsvisionlanguagemodels}, GLM-4.1V-9B-Thinking~\citep{vteam2025glm45vglm41vthinkingversatilemultimodal}, InternVL-Chat-V1-5~\citep{chen2024internvl}, and Qwen2.5-Omni-7B~\citep{xu2025qwen25omnitechnicalreport}. 
Kimi-VL-A3B-Instruct, DeepSeek-VL-Small, and InternVL-Chat-V1-5 are evaluated in English only; Qwen2.5-VL-32B-Instruct, Llama-3.2-11B-Vision-Instruct, and GLM-4.1V-9B-Thinking are evaluated in multiple languages. 
Qwen2.5-Omni-7B is evaluated across image, text, and speech in a multilingual setting.

\subsection{Prompt Templates and Experimental Settings}
\label{sec:testing prompt}

\paragraph{Prompt templates for evaluation tasks across modalities.}

Table \ref{prompts for English} presents the English prompts used in our evaluations across different modalities. When the query switches to another language, the corresponding translated version of the prompt will be used to ensure input-language consistency. Table \ref{tab:template for API} presents the prompts used to invoke closed-source model APIs for translation, answer-consistency evaluation, and answer extraction.

\begin{table}[htbp]
\footnotesize
\centering
\caption{English prompts for different modals.}
\label{prompts for English}

\begin{tabularx}{\textwidth}{@{} >{\RaggedRight\arraybackslash}p{3cm}| X @{}} 
\toprule
\textbf{Modal} & \textbf{Prompt Template} \\
\midrule
Text-Only/Rephrase  & Please answer the following culture-related question.\textbackslash n\{question\}\textbackslash n\{options\}\textbackslash nThis is a multiple-choice question. Please first return all possible option letters, then explain your choice in English. \\
\hline
VQA & Based on the image, please answer the following culture-related question.\textbackslash n\{question\}\textbackslash n\{options\}\textbackslash nThis is a multiple-choice question. Please first return all possible option letters, then explain your choice in English. \\
\hline
Speech question \& text options  & This is a culture-related question.\textbackslash n Based on the question mentioned in this audio, please choose the correct answers from the following provided options. \{options\}\textbackslash nThis is a multiple-choice question. Please first return all possible option letters, then explain your choice in English. \\
\hline
Speech question \& options & This is a culture-related question. Based on the question and options mentioned in this audio, please choose the correct options. This is a multiple-choice question. Please first return all possible option letters, then explain your choice in English.
\\
\bottomrule
\end{tabularx}
\end{table}

\paragraph{Experiments setting.}

\noindent All evaluations in this work are conducted in the zero-shot setting, using single-turn inference for each model on an NVIDIA H100 80G. For image inputs that exceed a model’s maximum allowable resolution, we proportionally downscale the image until it is under 1\~megapixel before testing. We decode with greedy search (no sampling; \texttt{do\_sample=false}, \texttt{num\_beams=1}), so temperature/top-\emph{p}/top-\emph{k} are not used; the maximum output length is set to 2048 tokens to ensure reproducibility. GPT-4o and Gemini 2.5 Flash are accessed via OpenRouter API platform, and Claude via the Anthropic API. For all closed-source models, we set \texttt{temperature=0} to minimize randomness and improve reproducibility. Our speech inputs are no longer than 30 seconds and sampled at 16 kHz—well below Gemini 2.5 Flash’s maximum speech-input duration and Qwen2.5-Omni-7B’s maximum input token limit. So we do not perform any input-length processing.

\begin{table}[!htbp]
\centering
\caption{Prompts Used with Closed-Source APIs for Translation, Answer Extraction, and Answer–Reference Consistency Checking
}
\label{tab:template for API}

\begin{tabularx}{\textwidth}{@{} l L @{}}
\toprule
\textbf{Task} & \textbf{Prompt} \\
\midrule

Translation & 
Translate the following sentence into English. \textbackslash n\{Input sentence\}\textbackslash nThis is a multiple-choice question in the cultural domain of \{Country\}. Pay particular attention to the terms in the input and use their official translations; if no official translation exists, you may use transliteration. Ensure accuracy, faithfulness, and fluency. Return only the translation; do not include any additional hints or analysis. \\
\midrule

LLM-as-Judge & 
You are an expert evaluator. Your task is to determine if the 'Model Answer' correctly and completely incorporates the information from the 'Knowledge Point'. 
Knowledge Point:\textbackslash n\{knowledge\_point\}\textbackslash nModel Answer\textbackslash n\{model\_answer\}\textbackslash n\textbackslash n. 
If the Model Answer correctly and completely includes the information from the Knowledge Point, or if the model's response aligns with the Knowledge Point's content and viewpoint. And the Model Answer doesn't contain any factual error, Answer 'yes'. Otherwise Answer 'no'. Please only return 'yes' or 'no'. \\
\midrule

Answer Extraction & 
This is a model's response to a multiple-choice question. First, understand the text, then extract the model's chosen options, returning only the option letters (e.g., A, B, C). Do not include the option content. Output the result in a format like [A, B]. If the response does not contain any final choice, return NA. \\

\bottomrule
\end{tabularx}
\end{table}

\subsection{Performance of LLMs on MMA-Asia across modalities}
\label{sec:Performance of LLMs}

The exact data corresponding to the bar chart in Section \ref{sec:cultural Awareness Disparity}, Figure \ref{fig:all} are presented in Tables \ref{tab:text-only}, \ref{tab:VQA}, \ref{tab:rephrase_vqa}, \ref{tab:speech_question_only}.

\begin{table*}[h!]
\centering
\small
\caption{Text-only modality performance with exact numbers, measured by Accuracy (\%): the number of items where the model’s choice exactly matches the correct option, divided by 500. ``-'' means ``not support''. The better-performing result among different languages within the same country is \textbf{bolded}.}
\label{tab:text-only}
\setlength{\tabcolsep}{5pt}

\sbox{\myfirsttablebox}{%
    \begin{tabular}{l|ccccccccc}
        \hline
        \textbf{Model} & \textbf{CN-zh} & \textbf{CN-en} & \textbf{ID-id} & \textbf{ID-en} & \textbf{IN-hi} & \textbf{IN-en} & \textbf{JP-ja} & \textbf{JP-en} & \textbf{KR-ko} \\
        \hline
        Gemini 2.5 Flash & \textbf{53.6} & 45.4 & \textbf{56.2} & 53.2 & 73.0 & \textbf{73.6} & 47.4 & \textbf{48.2} & \textbf{72.2} \\
        Claude Sonnet 4 & \textbf{53.8} & 53.6 & \textbf{55.6} & 54.8 & \textbf{74.4} & 74.2 & 48.8 & \textbf{50.6} & \textbf{71.4} \\
        GPT-4o & 36.0 & \textbf{50.2} & 53.6 & \textbf{55.8} & 73.2 & \textbf{77.2} & 49.8 & \textbf{53.2} & \textbf{71.4} \\
        GLM-4-9B-chat & \textbf{36.0} & 32.6 & 37.6 & \textbf{49.4} & 45.6 & \textbf{57.0} & 30.2 & \textbf{33.6} & \textbf{45.4} \\
        Mistral-8B-Instruct & 23.2 & \textbf{26.0} & 31.8 & \textbf{35.6} & 42.2 & \textbf{49.0} & 23.6 & \textbf{30.2} & \textbf{44.8} \\
        Baichuan-M2-32B & \textbf{51.2} & 49.8 & 50.6 & \textbf{53.2} & 54.6 & \textbf{69.0} & 39.6 & \textbf{40.8} & \textbf{56.4} \\
        Qwen3-30B-A3B... & \textbf{47.0} & 45.8 & 44.8 & \textbf{46.2} & 58.6 & \textbf{62.2} & \textbf{38.4} & 38.0 & 43.4 \\
        \hline
    \end{tabular}%
}
\sbox{\mysecondtablebox}{%
    \begin{tabular}{l|ccccccccc}
        \hline
        \textbf{Model} & \textbf{KR-en} & \textbf{MN-mn} & \textbf{MN-en} & \textbf{SG-zh} & \textbf{SG-en} & \textbf{SG-ms} & \textbf{SG-ta} & \textbf{VN-vi} & \textbf{VN-en} \\
        \hline
        Gemini 2.5 Flash & 65.2 & 58.2 & \textbf{70.0} & 34.0 & \textbf{43.4} & 39.0 & 31.4 & \textbf{73.6} & 73.0 \\
        Claude Sonnet 4 & 66.0 & 59.4 & \textbf{71.0} & 29.8 & \textbf{41.8} & 37.8 & 31.6 & \textbf{74.4} & 72.2 \\
        GPT-4o & 68.8 & 55.4 & \textbf{69.4} & 22.4 & \textbf{47.6} & 39.4 & 33.2 & \textbf{74.8} & 71.8 \\
        GLM-4-9B-chat & 42.8 & 23.0 & \textbf{52.6} & 20.2 & \textbf{24.8} & 18.4 & 19.4 & 47.6 & \textbf{51.0} \\
        Mistral-8B-Instruct & 40.8 & 13.0 & \textbf{50.8} & 17.2 & \textbf{25.2} & 22.4 & 15.4 & 41.8 & \textbf{48.6} \\
        Baichuan-M2-32B & 51.6 & 16.8 & \textbf{61.6} & 26.6 & \textbf{33.6} & 28.8 & 19.4 & \textbf{67.6} & 62.8 \\
        Qwen3-30B-A3B... & \textbf{49.8} & 44.6 & \textbf{65.0} & 26.4 & \textbf{29.2} & 23.6 & 22.8 & 57.8 & \textbf{65.8} \\
        \hline
    \end{tabular}%
}

\ifdim\wd\myfirsttablebox>\wd\mysecondtablebox
    \setlength{\maxtablewidth}{\wd\myfirsttablebox}
\else
    \setlength{\maxtablewidth}{\wd\mysecondtablebox}
\fi

\begin{tabular*}{\maxtablewidth}{l|@{\extracolsep{\fill}}ccccccccc}
\hline
\textbf{Model} & \textbf{CN-zh} & \textbf{CN-en} & \textbf{ID-id} & \textbf{ID-en} & \textbf{IN-hi} & \textbf{IN-en} & \textbf{JP-ja} & \textbf{JP-en} & \textbf{KR-ko} \\
\hline
Gemini 2.5 Flash & \textbf{53.6} & 45.4 & \textbf{56.2} & 53.2 & 73.0 & \textbf{73.6} & 47.4 & \textbf{48.2} & \textbf{72.2} \\
Claude Sonnet 4 & \textbf{53.8} & 53.6 & \textbf{55.6} & 54.8 & \textbf{74.4} & 74.2 & 48.8 & \textbf{50.6} & \textbf{71.4} \\
GPT-4o & 36.0 & \textbf{50.2} & 53.6 & \textbf{55.8} & 73.2 & \textbf{77.2} & 49.8 & \textbf{53.2} & \textbf{71.4} \\
GLM-4-9B-chat & \textbf{36.0} & 32.6 & 37.6 & \textbf{49.4} & 45.6 & \textbf{57.0} & 30.2 & \textbf{33.6} & \textbf{45.4} \\
Mistral-8B-Instruct & 23.2 & \textbf{26.0} & 31.8 & \textbf{35.6} & 42.2 & \textbf{49.0} & 23.6 & \textbf{30.2} & \textbf{44.8} \\
Baichuan-M2-32B & \textbf{51.2} & 49.8 & 50.6 & \textbf{53.2} & 54.6 & \textbf{69.0} & 39.6 & \textbf{40.8} & \textbf{56.4} \\
Qwen3-30B-A3B & \textbf{47.0} & 45.8 & 44.8 & \textbf{46.2} & 58.6 & \textbf{62.2} & \textbf{38.4} & 38.0 & 43.4 \\
\hline
\end{tabular*}

\vspace{0em}

\begin{tabular*}{\maxtablewidth}{l|@{\extracolsep{\fill}}ccccccccc}
\hline
\textbf{Model} & \textbf{KR-en} & \textbf{MN-mn} & \textbf{MN-en} & \textbf{SG-zh} & \textbf{SG-en} & \textbf{SG-ms} & \textbf{SG-ta} & \textbf{VN-vi} & \textbf{VN-en} \\
\hline
Gemini 2.5 Flash & 65.2 & 58.2 & \textbf{70.0} & 34.0 & \textbf{43.4} & 39.0 & 31.4 & \textbf{73.6} & 73.0 \\
Claude Sonnet 4 & 66.0 & 59.4 & \textbf{71.0} & 29.8 & \textbf{41.8} & 37.8 & 31.6 & \textbf{74.4} & 72.2 \\
GPT-4o & 68.8 & 55.4 & \textbf{69.4} & 22.4 & \textbf{47.6} & 39.4 & 33.2 & \textbf{74.8} & 71.8 \\
GLM-4-9B-chat & 42.8 & 23.0 & \textbf{52.6} & 20.2 & \textbf{24.8} & 18.4 & 19.4 & 47.6 & \textbf{51.0} \\
Mistral-8B-Instruct & 40.8 & 13.0 & \textbf{50.8} & 17.2 & \textbf{25.2} & 22.4 & 15.4 & 41.8 & \textbf{48.6} \\
Baichuan-M2-32B & 51.6 & 16.8 & \textbf{61.6} & 26.6 & \textbf{33.6} & 28.8 & 19.4 & \textbf{67.6} & 62.8 \\
Qwen3-30B-A3B & \textbf{49.8} & 44.6 & \textbf{65.0} & 26.4 & \textbf{29.2} & 23.6 & 22.8 & 57.8 & \textbf{65.8} \\
\hline
\end{tabular*}

\end{table*}

\begin{table*}[h!]
\centering
\footnotesize
\caption{VQA modality performance with exact numbers, measured by Accuracy (\%): the number of items where the model’s choice exactly matches the correct option, divided by 500. ``-'' means ``not support''. The better-performing result among different languages within the same country is bolded.}
\label{tab:VQA}
\setlength{\tabcolsep}{4pt}

\sbox{\myfirsttablebox}{%
    \begin{tabular}{l|ccccccccc}
        \hline
        \textbf{Model} & \textbf{CN-zh} & \textbf{CN-en} & \textbf{ID-id} & \textbf{ID-en} & \textbf{IN-hi} & \textbf{IN-en} & \textbf{JP-ja} & \textbf{JP-en} & \textbf{KR-ko} \\
        \hline
        Gemini 2.5 Flash & \textbf{45.6} & 38.8 & 48.6 & \textbf{49.8} & \textbf{65.4} & 42.4 & 54.2 & \textbf{58.0} & 72.0 \\
        Claude Sonnet 4 & \textbf{47.6} & 33.6 & \textbf{49.8} & 46.4 & \textbf{54.0} & 39.0 & 47.4 & \textbf{53.6} & 71.8 \\
        GPT-4o & 32.8 & \textbf{38.4} & \textbf{53.6} & 50.2 & \textbf{65.8} & 45.6 & 59.4 & \textbf{60.0} & \textbf{69.2} \\
        Qwen2.5-VL-32B... & \textbf{40.2} & 30.8 & 37.6 & \textbf{37.8} & 40.2 & \textbf{45.0} & \textbf{47.6} & 46.8 & 63.2 \\
        Llama-3.2-11B-Vision... & \textbf{21.6} & 15.2 & 26.2 & \textbf{31.2} & 35.4 & \textbf{40.2} & 27.2 & \textbf{35.2} & 30.0 \\
        Kimi-VL-A3B-instruct & - & \textbf{21.2} & - & \textbf{25.4} & - & \textbf{34.4} & - & \textbf{37.2} & - \\
        Qwen2.5-Omni-7B & \textbf{30.6} & 23.6 & 21.8 & \textbf{32.4} & 19.4 & \textbf{33.6} & 25.6 & \textbf{39.2} & 18.6 \\
        Deepseek-VL-Small & - & \textbf{12.2} & - & \textbf{25.4} & - & \textbf{25.0} & - & \textbf{26.0} & - \\
        GLM-4.1V-9B-Thinking & \textbf{46.4} & 26.2 & 28.6 & \textbf{34.8} & \textbf{50.8} & 43.4 & 44.0 & \textbf{44.8} & 61.0 \\
        InternVL-Chat-V1-5 & - & \textbf{15.4} & - & \textbf{23.0} & - & \textbf{39.2} & - & \textbf{40.6} & - \\
        \hline
    \end{tabular}%
}
\sbox{\mysecondtablebox}{%
    \begin{tabular}{l|ccccccccc}
        \hline
        \textbf{Model} & \textbf{KR-en} & \textbf{MN-mn} & \textbf{MN-en} & \textbf{SG-zh} & \textbf{SG-en} & \textbf{SG-ms} & \textbf{SG-ta} & \textbf{VN-vi} & \textbf{VN-en} \\
        \hline
        Gemini 2.5 Flash & \textbf{72.8} & 41.2 & \textbf{49.2} & 53.0 & \textbf{62.2} & 59.8 & 60.8 & \textbf{76.6} & 65.4 \\
        Claude Sonnet 4 & \textbf{74.4} & 35.0 & \textbf{49.0} & 34.6 & \textbf{57.2} & 55.2 & 49.8 & \textbf{73.2} & 67.4 \\
        GPT-4o & 65.8 & 41.4 & \textbf{53.0} & 30.2 & \textbf{70.6} & 68.4 & 62.4 & \textbf{75.8} & 63.8 \\
        Qwen2.5-VL-32B... & \textbf{67.0} & 20.6 & \textbf{45.2} & 40.8 & \textbf{52.8} & 45.6 & 22.6 & \textbf{65.0} & 64.6 \\
        Llama-3.2-11B-Vision... & \textbf{47.8} & 5.8 & \textbf{21.6} & 32.2 & \textbf{39.4} & 35.0 & 15.0 & 48.6 & \textbf{49.6} \\
        Kimi-VL-A3B-instruct & \textbf{56.0} & - & - & 38.6 & \textbf{43.6} & - & - & - & \textbf{50.0} \\
        Qwen2.5-Omni-7B & \textbf{49.2} & 3.2 & \textbf{36.6} & 30.0 & \textbf{43.4} & 38.6 & 14.2 & \textbf{59.8} & 58.2 \\
        Deepseek-VL-Small & \textbf{30.0} & - & - & 16.4 & \textbf{27.2} & - & - & - & \textbf{41.6} \\
        GLM-4.1V-9B-Thinking & \textbf{65.4} & 28.0 & \textbf{39.4} & 44.8 & \textbf{52.8} & 41.8 & 34.8 & 59.4 & \textbf{59.8} \\
        InternVL-Chat-V1-5 & \textbf{56.6} & - & - & 28.0 & \textbf{44.8} & - & - & - & \textbf{54.0} \\
        \hline
    \end{tabular}%
}

\ifdim\wd\myfirsttablebox>\wd\mysecondtablebox
    \setlength{\maxtablewidth}{\wd\myfirsttablebox}
\else
    \setlength{\maxtablewidth}{\wd\mysecondtablebox}
\fi

\begin{tabular*}{\maxtablewidth}{l|@{\extracolsep{\fill}}ccccccccc}
\hline
\textbf{Model} & \textbf{CN-zh} & \textbf{CN-en} & \textbf{ID-id} & \textbf{ID-en} & \textbf{IN-hi} & \textbf{IN-en} & \textbf{JP-ja} & \textbf{JP-en} & \textbf{KR-ko} \\
\hline
Gemini 2.5 Flash & \textbf{45.6} & 38.8 & 48.6 & \textbf{49.8} & \textbf{65.4} & 42.4 & 54.2 & \textbf{58.0} & 72.0 \\
Claude Sonnet 4 & \textbf{47.6} & 33.6 & \textbf{49.8} & 46.4 & \textbf{54.0} & 39.0 & 47.4 & \textbf{53.6} & 71.8 \\
GPT-4o & 32.8 & \textbf{38.4} & \textbf{53.6} & 50.2 & \textbf{65.8} & 45.6 & 59.4 & \textbf{60.0} & \textbf{69.2} \\
Qwen2.5-VL-32B & \textbf{40.2} & 30.8 & 37.6 & \textbf{37.8} & 40.2 & \textbf{45.0} & \textbf{47.6} & 46.8 & 63.2 \\
Llama-3.2-11B-Vision & \textbf{21.6} & 15.2 & 26.2 & \textbf{31.2} & 35.4 & \textbf{40.2} & 27.2 & \textbf{35.2} & 30.0 \\
Kimi-VL-A3B-instruct & - & \textbf{21.2} & - & \textbf{25.4} & - & \textbf{34.4} & - & \textbf{37.2} & - \\
Qwen2.5-Omni-7B & \textbf{30.6} & 23.6 & 21.8 & \textbf{32.4} & 19.4 & \textbf{33.6} & 25.6 & \textbf{39.2} & 18.6 \\
Deepseek-VL-Small & - & \textbf{12.2} & - & \textbf{25.4} & - & \textbf{25.0} & - & \textbf{26.0} & - \\
GLM-4.1V-9B-Thinking & \textbf{46.4} & 26.2 & 28.6 & \textbf{34.8} & \textbf{50.8} & 43.4 & 44.0 & \textbf{44.8} & 61.0 \\
InternVL-Chat-V1-5 & - & \textbf{15.4} & - & \textbf{23.0} & - & \textbf{39.2} & - & \textbf{40.6} & - \\
\hline
\end{tabular*}

\vspace{0em} 

\begin{tabular*}{\maxtablewidth}{l|@{\extracolsep{\fill}}ccccccccc}
\hline
\textbf{Model} & \textbf{KR-en} & \textbf{MN-mn} & \textbf{MN-en} & \textbf{SG-zh} & \textbf{SG-en} & \textbf{SG-ms} & \textbf{SG-ta} & \textbf{VN-vi} & \textbf{VN-en} \\
\hline
Gemini 2.5 Flash & \textbf{72.8} & 41.2 & \textbf{49.2} & 53.0 & \textbf{62.2} & 59.8 & 60.8 & \textbf{76.6} & 65.4 \\
Claude Sonnet 4 & \textbf{74.4} & 35.0 & \textbf{49.0} & 34.6 & \textbf{57.2} & 55.2 & 49.8 & \textbf{73.2} & 67.4 \\
GPT-4o & 65.8 & 41.4 & \textbf{53.0} & 30.2 & \textbf{70.6} & 68.4 & 62.4 & \textbf{75.8} & 63.8 \\
Qwen2.5-VL-32B & \textbf{67.0} & 20.6 & \textbf{45.2} & 40.8 & \textbf{52.8} & 45.6 & 22.6 & \textbf{65.0} & 64.6 \\
Llama-3.2-11B-Vision & \textbf{47.8} & 5.8 & \textbf{21.6} & 32.2 & \textbf{39.4} & 35.0 & 15.0 & 48.6 & \textbf{49.6} \\
Kimi-VL-A3B-instruct & \textbf{56.0} & - & - & 38.6 & \textbf{43.6} & - & - & - & \textbf{50.0} \\
Qwen2.5-Omni-7B & \textbf{49.2} & 3.2 & \textbf{36.6} & 30.0 & \textbf{43.4} & 38.6 & 14.2 & \textbf{59.8} & 58.2 \\
Deepseek-VL-Small & \textbf{30.0} & - & - & 16.4 & \textbf{27.2} & - & - & - & \textbf{41.6} \\
GLM-4.1V-9B-Thinking & \textbf{65.4} & 28.0 & \textbf{39.4} & 44.8 & \textbf{52.8} & 41.8 & 34.8 & 59.4 & \textbf{59.8} \\
InternVL-Chat-V1-5 & \textbf{56.6} & - & - & 28.0 & \textbf{44.8} & - & - & - & \textbf{54.0} \\
\hline
\end{tabular*}

\end{table*}

\begin{table*}[h!]
\centering
\footnotesize
\caption{Rephrase VQA (Text-Only) modality performance with exact numbers, measured by Accuracy (\%): the number of items where the model’s choice exactly matches the correct option, divided by 500. ``-'' means ``not support''. The better-performing result among different languages within the same country is \textbf{bolded}.}
\label{tab:rephrase_vqa}
\setlength{\tabcolsep}{4pt}

\sbox{\myfirsttablebox}{%
    \begin{tabular}{l|ccccccccc}
        \hline
        \textbf{Model} & \textbf{CN-zh} & \textbf{CN-en} & \textbf{ID-id} & \textbf{ID-en} & \textbf{IN-hi} & \textbf{IN-en} & \textbf{JP-ja} & \textbf{JP-en} & \textbf{KR-ko} \\
        \hline
        Gemini 2.5 Flash & \textbf{45.2} & 37.2 & \textbf{52.0} & 48.8 & \textbf{67.0} & 53.8 & 62.4 & \textbf{65.6} & \textbf{76.4} \\
        Claude Sonnet 4 & \textbf{52.8} & 48.0 & \textbf{53.8} & 52.6 & \textbf{68.6} & 51.8 & 67.8 & \textbf{68.2} & \textbf{79.6} \\
        GPT-4o & 41.0 & \textbf{46.2} & \textbf{57.8} & 55.2 & \textbf{69.0} & 59.6 & \textbf{67.0} & 66.4 & 73.2 \\
        Qwen2.5-VL-32B... & \textbf{45.0} & 34.4 & \textbf{45.0} & 44.4 & \textbf{51.1} & 47.0 & 56.0 & \textbf{58.4} & 66.8 \\
        Llama-3.2-11B-Vision... & \textbf{18.4} & 15.6 & \textbf{35.2} & 30.4 & \textbf{39.8} & 36.6 & 39.2 & \textbf{45.6} & 43.8 \\
        Kimi-VL-A3B-Instruct & - & \textbf{21.6} & - & \textbf{24.6} & - & \textbf{32.2} & - & \textbf{43.8} & - \\
        Qwen2.5-Omni-7B & \textbf{31.4} & 22.8 & 23.4 & \textbf{35.8} & 28.4 & \textbf{42.8} & 35.0 & \textbf{51.4} & 29.6 \\
        Deepseek-VL-Small & - & \textbf{15.6} & - & \textbf{30.6} & - & \textbf{29.2} & - & \textbf{40.4} & - \\
        GLM-4.1V-9B-Thinking & \textbf{46.4} & 30.6 & \textbf{37.6} & 33.6 & \textbf{51.4} & 33.8 & 49.0 & \textbf{53.0} & 66.6 \\
        InternVL-Chat-V1.5 & - & \textbf{16.8} & - & \textbf{27.0} & - & \textbf{37.4} & - & \textbf{40.9} & - \\
        \hline
    \end{tabular}%
}
\sbox{\mysecondtablebox}{%
    \begin{tabular}{l|ccccccccc}
        \hline
        \textbf{Model} & \textbf{KR-en} & \textbf{MN-mn} & \textbf{MN-en} & \textbf{SG-zh} & \textbf{SG-en} & \textbf{SG-ms} & \textbf{SG-ta} & \textbf{VN-vi} & \textbf{VN-en} \\
        \hline
        Gemini 2.5 Flash & 76.4 & 47.8 & \textbf{49.2} & 54.6 & \textbf{63.6} & 59.6 & 52.4 & \textbf{76.0} & 71.6 \\
        Claude Sonnet 4 & 77.0 & 38.8 & \textbf{56.0} & 46.0 & \textbf{65.2} & 61.0 & 51.8 & \textbf{76.0} & 73.6 \\
        GPT-4o & \textbf{75.2} & 43.2 & \textbf{59.8} & 30.2 & \textbf{68.2} & 66.4 & 53.6 & \textbf{79.4} & 72.8 \\
        Qwen2.5-VL-32B... & \textbf{72.0} & 19.0 & \textbf{56.2} & 36.4 & \textbf{57.6} & 50.8 & 25.4 & 69.2 & \textbf{70.0} \\
        Llama-3.2-11B-Vision... & \textbf{54.6} & 10.2 & \textbf{29.0} & 25.4 & \textbf{45.2} & 40.2 & 20.0 & 39.4 & \textbf{51.2} \\
        Kimi-VL-A3B-Instruct & \textbf{53.8} & - & - & \textbf{40.4} & 36.6 & - & - & - & \textbf{40.0} \\
        Qwen2.5-Omni-7B & \textbf{60.6} & 2.6 & \textbf{43.4} & 21.4 & \textbf{43.6} & 29.4 & 9.8 & 58.4 & \textbf{60.0} \\
        Deepseek-VL-Small & \textbf{58.8} & - & - & 25.2 & \textbf{39.2} & - & - & - & \textbf{58.4} \\
        GLM-4.1V-9B-Thinking & \textbf{67.0} & 26.6 & \textbf{43.8} & 42.0 & \textbf{51.0} & 40.4 & 34.8 & \textbf{66.2} & 62.8 \\
        InternVL-Chat-V1.5 & \textbf{48.6} & - & - & 36.0 & \textbf{41.6} & - & - & - & \textbf{46.0} \\
        \hline
    \end{tabular}%
}

\ifdim\wd\myfirsttablebox>\wd\mysecondtablebox
    \setlength{\maxtablewidth}{\wd\myfirsttablebox}
\else
    \setlength{\maxtablewidth}{\wd\mysecondtablebox}
\fi

\begin{tabular*}{\maxtablewidth}{l|@{\extracolsep{\fill}}ccccccccc}
\hline
\textbf{Model} & \textbf{CN-zh} & \textbf{CN-en} & \textbf{ID-id} & \textbf{ID-en} & \textbf{IN-hi} & \textbf{IN-en} & \textbf{JP-ja} & \textbf{JP-en} & \textbf{KR-ko} \\
\hline
Gemini 2.5 Flash & \textbf{45.2} & 37.2 & \textbf{52.0} & 48.8 & \textbf{67.0} & 53.8 & 62.4 & \textbf{65.6} & \textbf{76.4} \\
Claude Sonnet 4 & \textbf{52.8} & 48.0 & \textbf{53.8} & 52.6 & \textbf{68.6} & 51.8 & 67.8 & \textbf{68.2} & \textbf{79.6} \\
GPT-4o & 41.0 & \textbf{46.2} & \textbf{57.8} & 55.2 & \textbf{69.0} & 59.6 & \textbf{67.0} & 66.4 & 73.2 \\
Qwen2.5-VL-32B & \textbf{45.0} & 34.4 & \textbf{45.0} & 44.4 & \textbf{51.1} & 47.0 & 56.0 & \textbf{58.4} & 66.8 \\
Llama-3.2-11B-Vision & \textbf{18.4} & 15.6 & \textbf{35.2} & 30.4 & \textbf{39.8} & 36.6 & 39.2 & \textbf{45.6} & 43.8 \\
Kimi-VL-A3B-Instruct & - & \textbf{21.6} & - & \textbf{24.6} & - & \textbf{32.2} & - & \textbf{43.8} & - \\
Qwen2.5-Omni-7B & \textbf{31.4} & 22.8 & 23.4 & \textbf{35.8} & 28.4 & \textbf{42.8} & 35.0 & \textbf{51.4} & 29.6 \\
Deepseek-VL-Small & - & \textbf{15.6} & - & \textbf{30.6} & - & \textbf{29.2} & - & \textbf{40.4} & - \\
GLM-4.1V-9B-Thinking & \textbf{46.4} & 30.6 & \textbf{37.6} & 33.6 & \textbf{51.4} & 33.8 & 49.0 & \textbf{53.0} & 66.6 \\
InternVL-Chat-V1.5 & - & \textbf{16.8} & - & \textbf{27.0} & - & \textbf{37.4} & - & \textbf{40.9} & - \\
\hline
\end{tabular*}

\vspace{0em}

\begin{tabular*}{\maxtablewidth}{l|@{\extracolsep{\fill}}ccccccccc}
\hline
\textbf{Model} & \textbf{KR-en} & \textbf{MN-mn} & \textbf{MN-en} & \textbf{SG-zh} & \textbf{SG-en} & \textbf{SG-ms} & \textbf{SG-ta} & \textbf{VN-vi} & \textbf{VN-en} \\
\hline
Gemini 2.5 Flash & 76.4 & 47.8 & \textbf{49.2} & 54.6 & \textbf{63.6} & 59.6 & 52.4 & \textbf{76.0} & 71.6 \\
Claude Sonnet 4 & 77.0 & 38.8 & \textbf{56.0} & 46.0 & \textbf{65.2} & 61.0 & 51.8 & \textbf{76.0} & 73.6 \\
GPT-4o & \textbf{75.2} & 43.2 & \textbf{59.8} & 30.2 & \textbf{68.2} & 66.4 & 53.6 & \textbf{79.4} & 72.8 \\
Qwen2.5-VL-32B & \textbf{72.0} & 19.0 & \textbf{56.2} & 36.4 & \textbf{57.6} & 50.8 & 25.4 & 69.2 & \textbf{70.0} \\
Llama-3.2-11B-Vision & \textbf{54.6} & 10.2 & \textbf{29.0} & 25.4 & \textbf{45.2} & 40.2 & 20.0 & 39.4 & \textbf{51.2} \\
Kimi-VL-A3B-Instruct & \textbf{53.8} & - & - & \textbf{40.4} & 36.6 & - & - & - & \textbf{40.0} \\
Qwen2.5-Omni-7B & \textbf{60.6} & 2.6 & \textbf{43.4} & 21.4 & \textbf{43.6} & 29.4 & 9.8 & 58.4 & \textbf{60.0} \\
Deepseek-VL-Small & \textbf{58.8} & - & - & 25.2 & \textbf{39.2} & - & - & - & \textbf{58.4} \\
GLM-4.1V-9B-Thinking & \textbf{67.0} & 26.6 & \textbf{43.8} & 42.0 & \textbf{51.0} & 40.4 & 34.8 & \textbf{66.2} & 62.8 \\
InternVL-Chat-V1.5 & \textbf{48.6} & - & - & 36.0 & \textbf{41.6} & - & - & - & \textbf{46.0} \\
\hline
\end{tabular*}

\end{table*}

\begin{table*}[h!]
\centering
\footnotesize
\caption{Speech modality (speech question \& text options) performance with exact numbers, measured by Accuracy (\%): the number of items where the model’s choice exactly matches the correct option, divided by 500. ``-'' means ``not support''. The better-performing result among different languages within the same country is \textbf{bolded}.}
\label{tab:speech_question_only}
\setlength{\tabcolsep}{5pt}

\newsavebox{\mythirdtablebox}

\sbox{\myfirsttablebox}{%
    \begin{tabular}{l|c cc c cc c cc}
        \hline
        \textbf{Model} & \textbf{CN-zh} & \textbf{CN-en} & \textbf{CN-en} & \textbf{ID-id} & \textbf{ID-en} & \textbf{ID-en} & \textbf{IN-hi} & \textbf{IN-en} & \textbf{IN-en} \\
        \hline
         & - & Acc & NoAcc & - & Acc & NoAcc & - & Acc & NoAcc \\
        \hline
        Qwen2.5-Omni-7B & \textbf{26.2} & 21.2 & 21.0 & 29.4 & 29.4 & \textbf{32.0} & 33.2 & \textbf{33.2} & 30.4 \\
        Gemini 2.5 Flash & \textbf{40.6} & 37.6 & 36.0 & 36.6 & 40.2 & \textbf{40.4} & \textbf{53.4} & 25.4 & 24.2 \\
        \hline
    \end{tabular}%
}

\sbox{\mysecondtablebox}{%
    \begin{tabular}{l|c cc c cc c cc}
        \hline
        \textbf{Model} & \textbf{JP-ja} & \textbf{JP-en} & \textbf{JP-en} & \textbf{KR-ko} & \textbf{KR-en} & \textbf{KR-en} & \textbf{MN-mn} & \textbf{MN-en} & \textbf{MN-en} \\
        \hline
         & - & Acc & NoAcc & - & Acc & NoAcc & - & Acc & NoAcc \\
        \hline
        Qwen2.5-Omni-7B & \textbf{45.4} & 41.8 & - & \textbf{53.6} & 51.4 & - & 43.8 & \textbf{47.8} & - \\
        Gemini 2.5 Flash & 52.4 & \textbf{60.2} & 57.0 & \textbf{73.6} & 70.2 & 72.0 & 32.0 & 48.0 & \textbf{50.4} \\
        \hline
    \end{tabular}%
}

\sbox{\mythirdtablebox}{%
    \begin{tabular}{l|c cc c c c cc}
        \hline
        \textbf{Model} & \textbf{SG-zh} & \textbf{SG-en} & \textbf{SG-en} & \textbf{SG-ms} & \textbf{SG-ta} & \textbf{VN-vi} & \textbf{VN-en} & \textbf{VN-en} \\
        \hline
         & - & Acc & NoAcc & - & - & - & Acc & NoAcc \\
        \hline
        Qwen2.5-Omni-7B & \textbf{39.4} & 39.4 & 37.0 & - & - & - & \textbf{54.6} & 52.0 \\
        Gemini 2.5 Flash & 47.0 & \textbf{55.2} & 54.4 & 41.4 & 39.6 & 60.6 & \textbf{68.6} & 67.8 \\
        \hline
    \end{tabular}%
}

\setlength{\maxtablewidth}{\wd\myfirsttablebox}
\ifdim\wd\mysecondtablebox>\maxtablewidth\setlength{\maxtablewidth}{\wd\mysecondtablebox}\fi
\ifdim\wd\mythirdtablebox>\maxtablewidth\setlength{\maxtablewidth}{\wd\mythirdtablebox}\fi

\begin{tabular*}{\maxtablewidth}{l|@{\extracolsep{\fill}}c cc c cc c cc}
\hline
\textbf{Model} & \textbf{CN-zh} & \textbf{CN-en} & \textbf{CN-en} & \textbf{ID-id} & \textbf{ID-en} & \textbf{ID-en} & \textbf{IN-hi} & \textbf{IN-en} & \textbf{IN-en} \\
\hline
 & - & Acc & NoAcc & - & Acc & NoAcc & - & Acc & NoAcc \\
\hline
Qwen2.5-Omni-7B & \textbf{26.2} & 21.2 & 21.0 & 29.4 & 29.4 & \textbf{32.0} & 33.2 & \textbf{33.2} & 30.4 \\
Gemini 2.5 Flash & \textbf{40.6} & 37.6 & 36.0 & 36.6 & 40.2 & \textbf{40.4} & \textbf{53.4} & 25.4 & 24.2 \\
\hline
\end{tabular*}

\vspace{0em} 

\begin{tabular*}{\maxtablewidth}{l|@{\extracolsep{\fill}}c cc c cc c cc}
\hline
\textbf{Model} & \textbf{JP-ja} & \textbf{JP-en} & \textbf{JP-en} & \textbf{KR-ko} & \textbf{KR-en} & \textbf{KR-en} & \textbf{MN-mn} & \textbf{MN-en} & \textbf{MN-en} \\
\hline
 & - & Acc & NoAcc & - & Acc & NoAcc & - & Acc & NoAcc \\
\hline
Qwen2.5-Omni-7B & \textbf{45.4} & 41.8 & - & \textbf{53.6} & 51.4 & - & 43.8 & \textbf{47.8} & - \\
Gemini 2.5 Flash & 52.4 & \textbf{60.2} & 57.0 & \textbf{73.6} & 70.2 & 72.0 & 32.0 & 48.0 & \textbf{50.4} \\
\hline
\end{tabular*}

\vspace{0em}

\begin{tabular*}{\maxtablewidth}{l|@{\extracolsep{\fill}}c cc c c c cc}
\hline
\textbf{Model} & \textbf{SG-zh} & \textbf{SG-en} & \textbf{SG-en} & \textbf{SG-ms} & \textbf{SG-ta} & \textbf{VN-vi} & \textbf{VN-en} & \textbf{VN-en} \\
\hline
 & - & Acc & NoAcc & - & - & - & Acc & NoAcc \\
\hline
Qwen2.5-Omni-7B & \textbf{39.4} & 39.4 & 37.0 & - & - & - & \textbf{54.6} & 52.0 \\
Gemini 2.5 Flash & 47.0 & \textbf{55.2} & 54.4 & 41.4 & 39.6 & 60.6 & \textbf{68.6} & 67.8 \\
\hline
\end{tabular*}

\end{table*}

\subsection{Results for fully spoken question and answering}
\label{sec:results for fully spoken}

\begin{figure*}[h]
  \centering
  \includegraphics[trim=0 0 0 0, clip,width=\textwidth]{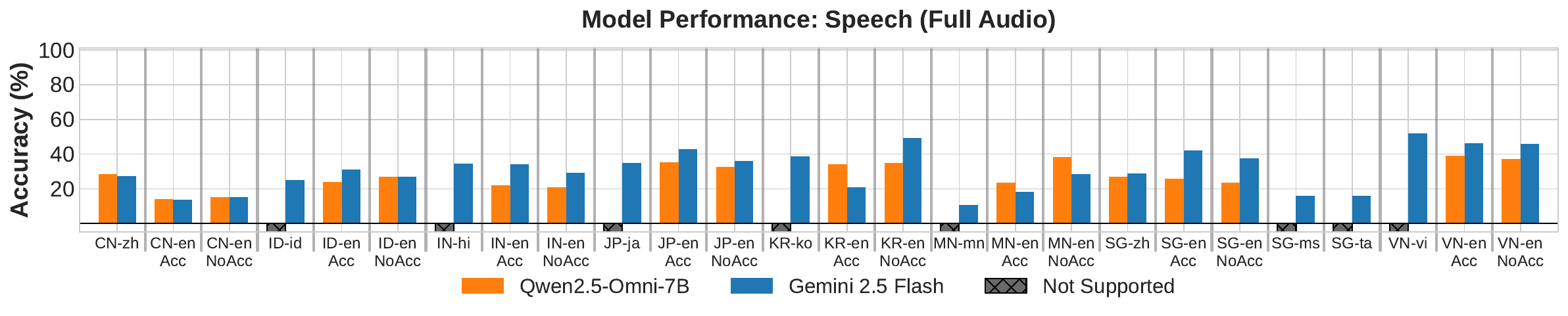}
  \captionsetup{skip=2pt}
  \caption{Speech modality (speech question \& options) performance with exact numbers, measured by Accuracy (\%): the number of items where the model’s choice exactly matches the correct option, divided by 500.}
  \label{fig:result for fully spoken}
\end{figure*}

We considered two configurations when constructing the TTS-Spoken QA dataset: (i) converting only the question stem to speech while keeping the answer options as text, and (ii) converting both the stem and the options to speech. To preserve comparability with VQA under controlled variables and to minimize ambiguity introduced by fully spoken options, our main experiments adopt the “spoken stem + textual options” setting across five evaluation dimensions. Results for the fully spoken setting (spoken stem and spoken options) on the test set are provided in Figure \ref{fig:result for fully spoken} and Table \ref{tab:fully speech} for reference. We find that converting both the question and options to speech leads to a significant performance drop compared with the “spoken question + textual options” configuration, indicating that spoken options introduce greater uncertainty than the spoken question itself. This warrants further investigation in future work.

\begin{table*}[h!]
\centering
\footnotesize
\caption{Speech modality (speech question \& options) performance with exact numbers, measured by Accuracy (\%): the number of items where the model’s choice exactly matches the correct option, divided by 500. ``-'' means ``not support''. The better-performing result among different languages within the same country is \textbf{bolded}.}
\label{tab:fully speech}
\setlength{\tabcolsep}{5pt}

\sbox{\myfirsttablebox}{%
    \begin{tabular}{l|c cc c cc c cc}
        \hline
        \textbf{Model} & \textbf{CN-zh} & \textbf{CN-en} & \textbf{CN-en} & \textbf{ID-id} & \textbf{ID-en} & \textbf{ID-en} & \textbf{IN-hi} & \textbf{IN-en} & \textbf{IN-en} \\
        \hline
         & - & Acc & NoAcc & - & Acc & NoAcc & - & Acc & NoAcc \\
        \hline
        Qwen2.5-Omni-7B & \textbf{28.4} & 14.2 & 15.2 & - & 23.8 & \textbf{27.0} & - & \textbf{22.2} & 20.8 \\
        Gemini 2.5 Flash & \textbf{27.4} & 13.8 & 15.2 & 25.0 & \textbf{31.0} & 27.0 & \textbf{34.6} & 34.2 & 29.4 \\
        \hline
    \end{tabular}%
}

\sbox{\mysecondtablebox}{%
    \begin{tabular}{l|c cc c cc c cc}
        \hline
        \textbf{Model} & \textbf{JP-ja} & \textbf{JP-en} & \textbf{JP-en} & \textbf{KR-ko} & \textbf{KR-en} & \textbf{KR-en} & \textbf{MN-mn} & \textbf{MN-en} & \textbf{MN-en} \\
        \hline
         & - & Acc & NoAcc & - & Acc & NoAcc & - & Acc & NoAcc \\
        \hline
        Qwen2.5-Omni-7B & - & \textbf{35.2} & 32.8 & - & 34.0 & \textbf{35.0} & - & 23.6 & \textbf{38.4} \\
        Gemini 2.5 Flash & 35.0 & \textbf{42.8} & 36.2 & 38.6 & 20.8 & \textbf{49.4} & 10.6 & 18.2 & \textbf{28.6} \\
        \hline
    \end{tabular}%
}

\sbox{\mythirdtablebox}{%
    \begin{tabular}{l|c cc c c c cc}
        \hline
        \textbf{Model} & \textbf{SG-zh} & \textbf{SG-en} & \textbf{SG-en} & \textbf{SG-ms} & \textbf{SG-ta} & \textbf{VN-vi} & \textbf{VN-en} & \textbf{VN-en} \\
        \hline
         & - & Acc & NoAcc & - & - & - & Acc & NoAcc \\
        \hline
        Qwen2.5-Omni-7B & \textbf{27.0} & 26.0 & 23.6 & - & - & - & \textbf{39.2} & 37.0 \\
        Gemini 2.5 Flash & 28.8 & \textbf{42.2} & 37.6 & 16.0 & 15.8 & \textbf{52.0} & 46.2 & 45.8 \\
        \hline
    \end{tabular}%
}

\setlength{\maxtablewidth}{\wd\myfirsttablebox}
\ifdim\wd\mysecondtablebox>\maxtablewidth\setlength{\maxtablewidth}{\wd\mysecondtablebox}\fi
\ifdim\wd\mythirdtablebox>\maxtablewidth\setlength{\maxtablewidth}{\wd\mythirdtablebox}\fi

\begin{tabular*}{\maxtablewidth}{l|@{\extracolsep{\fill}}c cc c cc c cc}
\hline
\textbf{Model} & \textbf{CN-zh} & \textbf{CN-en} & \textbf{CN-en} & \textbf{ID-id} & \textbf{ID-en} & \textbf{ID-en} & \textbf{IN-hi} & \textbf{IN-en} & \textbf{IN-en} \\
\hline
 & - & Acc & NoAcc & - & Acc & NoAcc & - & Acc & NoAcc \\
\hline
Qwen2.5-Omni-7B & \textbf{28.4} & 14.2 & 15.2 & - & 23.8 & \textbf{27.0} & - & \textbf{22.2} & 20.8 \\
Gemini 2.5 Flash & \textbf{27.4} & 13.8 & 15.2 & 25.0 & \textbf{31.0} & 27.0 & \textbf{34.6} & 34.2 & 29.4 \\
\hline
\end{tabular*}

\vspace{0em}

\begin{tabular*}{\maxtablewidth}{l|@{\extracolsep{\fill}}c cc c cc c cc}
\hline
\textbf{Model} & \textbf{JP-ja} & \textbf{JP-en} & \textbf{JP-en} & \textbf{KR-ko} & \textbf{KR-en} & \textbf{KR-en} & \textbf{MN-mn} & \textbf{MN-en} & \textbf{MN-en} \\
\hline
 & - & Acc & NoAcc & - & Acc & NoAcc & - & Acc & NoAcc \\
\hline
Qwen2.5-Omni-7B & - & \textbf{35.2} & 32.8 & - & 34.0 & \textbf{35.0} & - & 23.6 & \textbf{38.4} \\
Gemini 2.5 Flash & 35.0 & \textbf{42.8} & 36.2 & 38.6 & 20.8 & \textbf{49.4} & 10.6 & 18.2 & \textbf{28.6} \\
\hline
\end{tabular*}

\vspace{0em}

\begin{tabular*}{\maxtablewidth}{l|@{\extracolsep{\fill}}c cc c c c cc}
\hline
\textbf{Model} & \textbf{SG-zh} & \textbf{SG-en} & \textbf{SG-en} & \textbf{SG-ms} & \textbf{SG-ta} & \textbf{VN-vi} & \textbf{VN-en} & \textbf{VN-en} \\
\hline
 & - & Acc & NoAcc & - & - & - & Acc & NoAcc \\
\hline
Qwen2.5-Omni-7B & \textbf{27.0} & 26.0 & 23.6 & - & - & - & \textbf{39.2} & 37.0 \\
Gemini 2.5 Flash & 28.8 & \textbf{42.2} & 37.6 & 16.0 & 15.8 & \textbf{52.0} & 46.2 & 45.8 \\
\hline
\end{tabular*}

\end{table*}

\subsection{Vision-ablated Prefix Replay}
\label{sec: Vision-ablated Prefix Replay}

We found that visual content increases
reasoning hallucinations in models compared to text-only QA, despite accurate image recognition capabilities. To validate our hypothesis, we propose a ``Vision-ablated Prefix Replay'' (VPR) method. This method enables a model with parameters $\theta$ to first describe the image contents based on image $\mathbf{x}^{\text{img}}$ and text prompt $\mathbf{x}^{\text{text}}$. After this initial description, we structurally ablate the visual condition and fix the prefix $\hat{\mathbf{S}}_{1:n}$ for subsequent reasoning generation. This evaluates the marginal contribution of visual conditions to reasoning. Specifically, assuming the model completes image description within the first $n$ tokens, we remove visual conditions starting from the $(n+1)$-th token and use only the text prompt and generated tokens as prefix. The joint probability distribution of the subsequent sequence $\mathbf{S}_{>n} = (s_{n+1}, \dots, s_T)$ can be expressed as:
{%
\setlength{\abovedisplayskip}{2pt}        
\setlength{\abovedisplayshortskip}{0pt}   
\setlength{\belowdisplayskip}{2pt}        
\setlength{\belowdisplayshortskip}{2pt}   
\begin{equation}
\small
p_{\theta}\!\bigl(\mathbf{S}_{>n}\mid \mathbf{x}_{\text{text}},\, \varnothing,\, \hat{\mathbf{S}}_{1:n}\bigr)
= \prod_{t=n+1}^{T} p_{\theta}\!\bigl(s_{t}\mid \mathbf{x}_{\text{text}},\, \varnothing,\, \hat{\mathbf{S}}_{1:n},\, \mathbf{s}_{n+1:t-1}\bigr).
\label{eq:replay-factorization}
\end{equation}
}%

Previous work has explored related ideas. For example, \citet{zhang2025crossmodalinformationflowmultimodal} attempt to suppress visual leakage by blocking the attention paths to image-token positions during decoding. However, this cannot fully eliminate the influence of visual content: in a causal language model, the information of earlier image tokens is encoded into subsequent question tokens, so residual visual information remains even when attention to image tokens is blocked. In contrast, our method first elicits a textual description of the image, then removes the visual input and recomputes the representations of the prefix tokens, thereby purging visual information and enabling a more precise assessment of the image content’s contribution to the model’s reasoning process.

For the question in Figure \ref{fig:replay result}, the completed responses under different generation modes are provided in Table \ref{tab:model_outputs_replay}.

\begin{table}[h!]
\centering
\footnotesize
\caption{Model outputs across different modes for the same question.}
\label{tab:model_outputs_replay}

\setlength{\tabcolsep}{3pt}

\begin{tabularx}{\textwidth}{ >{\bfseries}l  >{\RaggedRight\arraybackslash}X }
\toprule

Mode & Answer \\
\midrule

\begin{tabular}[t]{@{}l@{}}Rephrase VQA \\ baseline\end{tabular} & Answer: B. Yan Liang. Explanation: In historical records and the classic Chinese novel \textit{Romance of the Three Kingdoms}, Guan Yu is known for his legendary bravery and martial prowess. One of his most famous feats occurred during the Battle of Hulao Gate (虎牢关之战) in 194 CE, which took place during the late Eastern Han dynasty. Here's the context: \textbf{Hua Xiong}: During the Battle of Hulao Gate, Hua Xiong was a general serving under Dong Zhuo, a powerful warlord who controlled the Han imperial court. Hua Xiong challenged the coalition forces led by Yuan Shu, Yuan Shao, and others. Guan Yu volunteered to fight Hua Xiong and defeated him in battle. This event is well-known as one of Guan Yu's early displays of martial skill. \textbf{Yan Liang and Wen Chou}: After defeating Hua Xiong, Guan Yu continued to distinguish himself in battle. He later encountered two other generals, Yan Liang and Wen Chou, who were also serving under Dong Zhuo. In the same battle, Guan Yu fought and killed both Yan Liang and Wen Chou. This event further solidified his reputation as a formidable warrior. \textbf{Yu Jin}: Yu Jin was a general who served under Cao Cao, a rival of Guan Yu's lord, Liu Bei. While Yu Jin and Guan Yu did face each other in battle, Yu Jin was not killed by Guan Yu. Instead, Yu Jin surrendered to Guan Yu during the Battle of Fancheng (樊城之战) in 219 CE, after being trapped by water flooding caused by Guan Yu. Why the Correct Answer is B (Yan Liang): Yan Liang was explicitly mentioned in historical records and the \textit{Romance of the Three Kingdoms} as one of the generals killed by Guan Yu during the Battle of Hulao Gate. The other options do not align with the historical context: Hua Xiong was defeated but not killed by Guan Yu in the same battle. Wen Chou was also killed by Guan Yu, but the question asks for a single correct answer, and Yan Liang is more commonly associated with this specific event. Yu Jin was not killed by Guan Yu; he surrendered instead. Thus, the correct answer is B. Yan Liang. \\
\midrule

VQA baseline & The image depicts a statue of Guan Yu, one of the most famous and revered generals from the Three Kingdoms period in Chinese history. Guan Yu is known for his loyalty, bravery, and righteousness, and he is often associated with the Shu Han kingdom. Key Points to Consider: Guan Yu's Achievements: Guan Yu is celebrated for his military prowess and is known for several significant victories during the Three Kingdoms era. Hua Xiong (华雄): This is a key figure in the story of the "Battle of Changban" (or "Battle of Changping"). According to historical records and the novel Romance of the Three Kingdoms, Guan Yu famously defeated Hua Xiong in battle. Hua Xiong was a general serving under Dong Zhuo, and his defeat marked a turning point in the coalition against Dong Zhuo. Analysis of the Options: A. Hua Xiong: This is correct. Hua Xiong was indeed killed by Guan Yu during the Battle of Changban. This event is well-documented in both historical records and the Romance of the Three Kingdoms. B. Yan Liang (颜良): While Guan Yu did kill Yan Liang, this occurred later in his career, during the Battle of Baima (White Horse). However, the question specifically asks about the person in the picture, and the most iconic victory associated with Guan Yu is his defeat of Hua Xiong. C. Wen Chou (文丑): Wen Chou was another general defeated by Guan Yu, but this happened during the battle of Xiapi, which came after the Battle of Changban. D. Yu Jin (于禁): Yu Jin was a general of Cao Wei who surrendered to Guan Yu during the Battle of Fancheng. He was not killed by Guan Yu. Final Answer: The person in the picture (Guan Yu) is most famously known for killing Hua Xiong. Therefore, the correct answer is: A. \\
\midrule

\begin{tabular}[t]{@{}l@{}}Vision-ablated\\Prefix Replay\end{tabular} & The image depicts a statue of Guan Yu, one of the most famous and revered generals from the Three Kingdoms period in Chinese history. Guan Yu is known for his loyalty, bravery, and martial prowess. Analysis of the question: We need to determine which general was killed by Guan Yu based on historical records. \textbf{Hua Xiong (华雄)}: Hua Xiong was a general serving under Dong Zhuo during the late Eastern Han dynasty. He was killed by Guan Yu in the Battle of Hulao Gate, as described in the Romance of the Three Kingdoms. This event is one of the most iconic moments in Guan Yu's early career. \textbf{Yan Liang (颜良)}: Yan Liang was a general serving under Yuan Shao. He was killed by Guan Yu during the Battle of Baima (白马之战) in 200 CE. This battle was part of the larger conflict between Cao Cao and Yuan Shao. \textbf{Wen Chou (文丑)}: Wen Chou was another general serving under Yuan Shao. He was also killed by Guan Yu shortly after the Battle of Baima, during the same campaign against Yuan Shao. \textbf{Yu Jin (于禁)}: Yu Jin was a general serving under Cao Cao. He was not killed by Guan Yu but rather surrendered to Guan Yu during the Battle of Fancheng in 219 CE when Guan Yu's forces surrounded him. Yu Jin later returned to Cao Cao's service after Guan Yu's defeat. Correct Answer: Based on the historical context and the question, the general who was killed by Guan Yu is Yan Liang. Final Answer: B. \\
\bottomrule
\end{tabularx}
\vspace{-0.3cm}
\end{table}

\subsection{Analysis of Speech as a Cultural Prior}
\label{sec: Speech as a Cultural Prior}

Compared to images and text, speech input introduces greater uncertainty through environmental noise, homophony, and accents—with accents closely tied to cultural context. Our research reveals that accents function beyond mere noise. Testing synthetic speech in standard English versus multiple national accents, we found Qwen and Gemini outperformed their standard English baselines in 6 and 5 country-specific cultural settings (Figure \ref{fig:all}), respectively. Notably, Qwen achieved 2.8\% and 3.6\% accuracy gains for Indonesian and Japanese accents (Table~\ref{tab:speech_question_only}). We attribute this to systematic co-occurrence of accented English with country-specific entities and contexts in training corpora, enabling accents to serve as cultural and lexical priors during inference. Our findings demonstrate that accents can function as valuable cultural cues rather than simply noise sources for model exploitation.

\subsection{Consistency Analysis and Hyperparameters Setting for LLM-as-Judge}
\label{sec:LLM-as-Judge}

\noindent Before adopting the LLM-as-judge paradigm, we conducted a small-scale study to assess human--model agreement and inter-model agreement.

\paragraph{Human--model agreement.}
We sampled 50 items from the dataset, each comprising a multi-step reasoning question and its decomposed sub-questions. Three annotators independently judged whether the model’s answer was semantically consistent with the gold answer for each (binary: \emph{yes}/\emph{no}). For each sub-question, the human judgment was determined by majority vote. We then queried \textbf{Claude Sonnet~4}, \textbf{GPT-4o}, and \textbf{Gemini~2.5~Flash} via API to obtain their judgments on the same items. An item was counted as \emph{consistent} for a model only if the model’s judgments for all sub-questions and the final question matched the human judgments. Results showed human--model agreement of \textbf{98\%} for GPT-4o, \textbf{98\%} for Claude, and \textbf{96\%} for Gemini. Considering cost, we selected \textbf{GPT-4o} as the primary judge.

\paragraph{Inter-model agreement.}
Given the binary nature of the task and the observed human--model agreement rates (98\%, 98\%, 96\%), the conservative lower-bound on inter-model agreement is \textbf{96\%}. We therefore conclude that a single, top-performing judge model suffices for our setting, and cross-model adjudication is unnecessary.

\subsection{Rationale Unfaithfulness Rates Across Rephrase VQA and Spoken QA}
\label{sec:RUR}

\begin{figure}[h]
  \centering
  \begin{subfigure}{\textwidth}
    \centering
    \includegraphics[trim=0 10 0 20, clip,width=\linewidth]{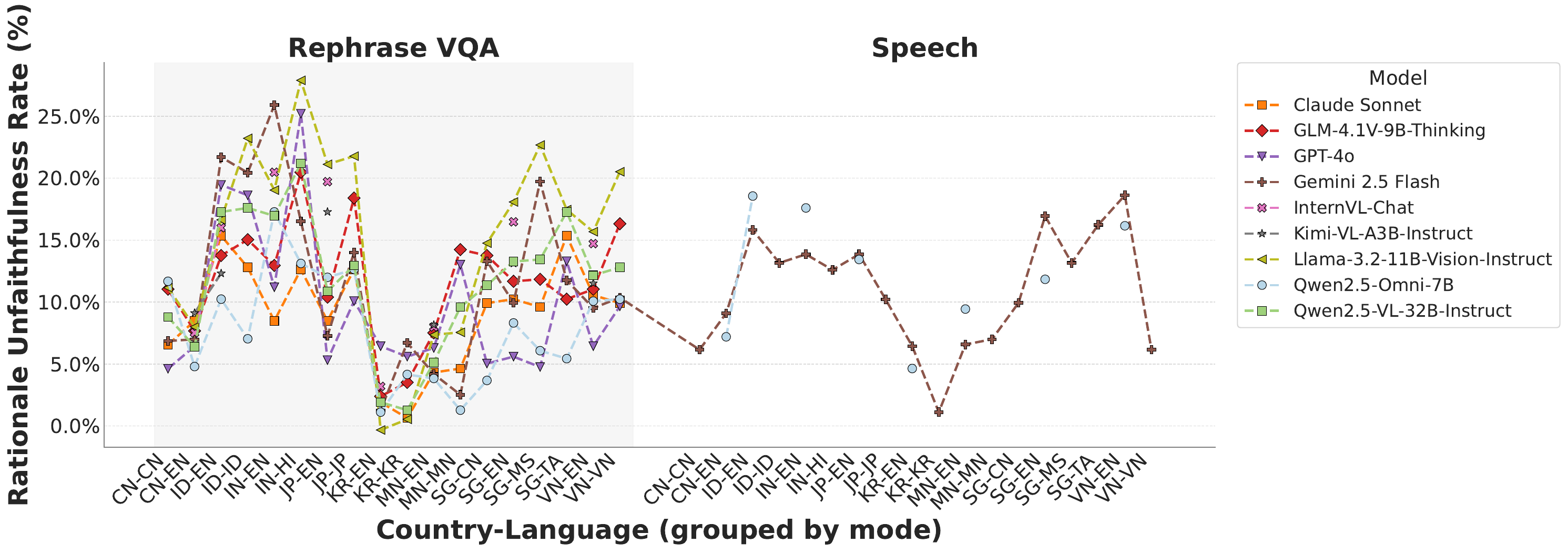}
    \label{fig:speech_and_rephrase}
  \end{subfigure}
  \captionsetup{skip=2pt}
  \caption{Rationale Unfaithfulness Rates of LLMs across Rephrase VQA and Speech.}
  \label{fig:RUR for Rephrase VQA and Speech}
\end{figure}

The results for LLMs’ Rationale Unfaithfulness Rates (RUR) across Rephrase VQA and Speech are shown in Fig. \ref{fig:RUR for Rephrase VQA and Speech}. We observe patterns consistent with the Text-Only and VQA modalities: closed-source models generally have lower RURs than open-source models, though they still fall within the 5\%–20\% range. Among open-source models, Llama shows a markedly higher RUR on non-Spanish languages than on Spanish, which we attribute to linguistic bias stemming from the disproportionately large share of Spanish in Llama’s training data relative to other languages.

\subsection{Attention heatmaps for incorrect model predictions}
\label{sec: Attention heatmaps}

\begin{figure*}[h]
  \centering
  \includegraphics[trim=0 170 0 0, clip,width=\textwidth]{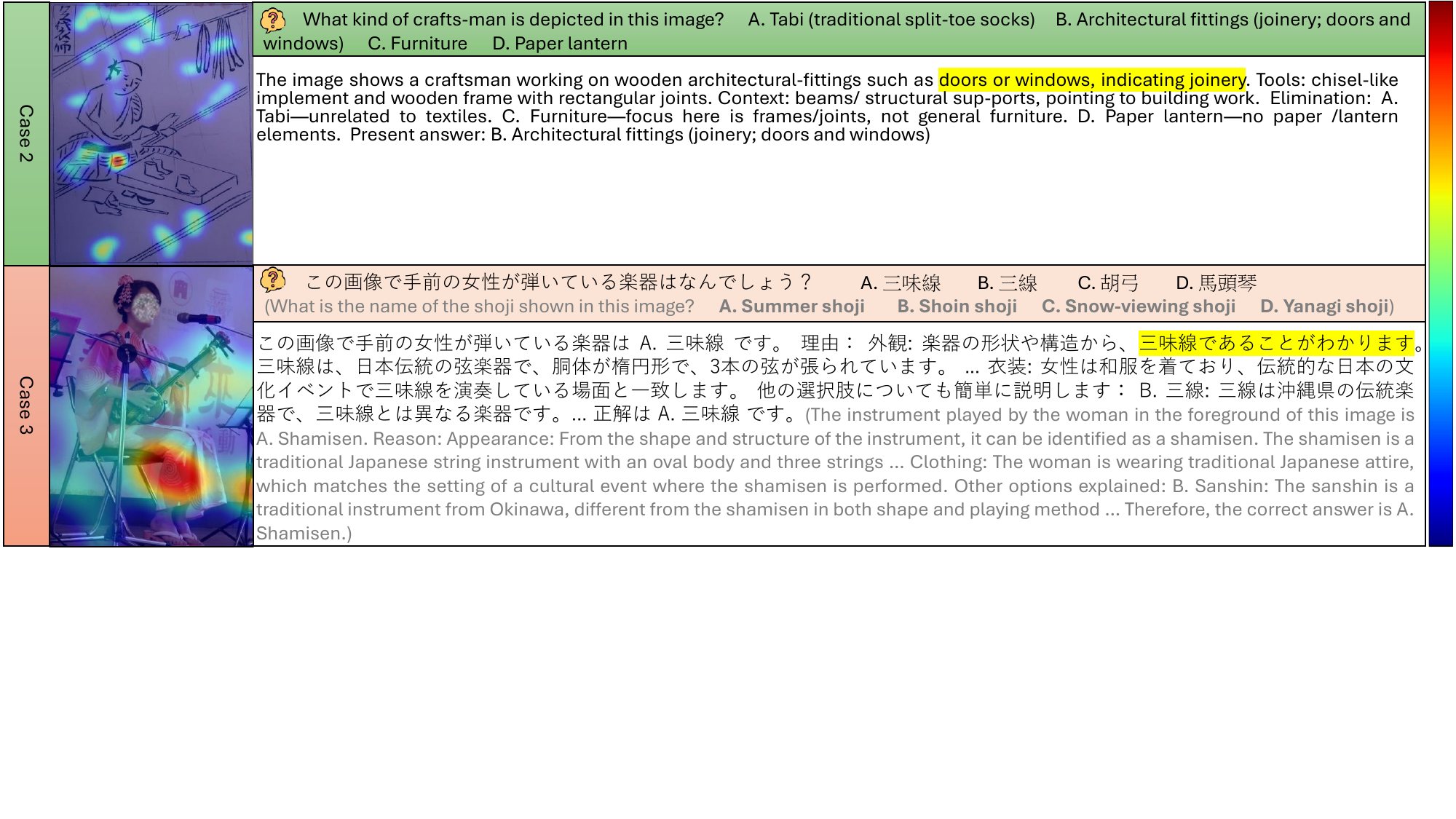}

  \caption{Attention heatmap visualization over image regions during incorrect model answers. Color scale from \BlueColorIcon blue (low) to \RedColorIcon red (high) indicates increasing model attention.}
  \label{fig:attention2}
\end{figure*}

Figures ~\ref{fig:attention2} shows attention heatmaps over image regions when the model produces incorrect answers. In Case 2, the model concentrates on the subject’s clothing and surrounding scene but overlooks the footwear beside the person and the key text in the upper-left corner that identifies the subject. Similarly, in Figure 3 the model focuses excessively on attire and playing posture while ignoring the crucial detail that the instrument’s body is wrapped in snakeskin. All of these support our finding: the model’s excessive focus on items mentioned in the prompt causes it to overlook critical cultural details.

\subsection{Sub-question Decomposition}
\label{Sec:Sub-question Decomposition}

Table \ref{table:multi-steps} illustrates examples of deconstructing multi-step reasoning questions into single-fact verification subquestions. We enumerate all single-fact subquestions embedded in each multi-step item to assess the model’s generalization.

\begin{table}[htbp]
\centering
\scriptsize
\caption{Sub-question decomposition of multi-step reasoning questions.}
\label{table:multi-steps}
\setlength{\tabcolsep}{4pt} 
\begin{tabularx}{\linewidth}{p{4.2cm} Y l} 
\toprule
\textbf{Multi-Step Question} & \textbf{Sub-Question} & \textbf{Answer for Sub-Question} \\
\midrule
\multirow{7}{4.0cm}{\RaggedRight Which of the following buildings belongs to the same ethnic characteristic architecture as the stilted building? A. Tulou B. Moxiaolou C. Diaofang D. Yaodong}
& What ethnic group is associated with the stilted building? & The Tujia ethnic group. \\
\cmidrule(l){2-3}
& What ethnic group is associated with the Tulou? & Hakka \\
\cmidrule(l){2-3}
& What ethnic group is associated with the Moxiaolou? & The Tujia ethnic group. \\
\cmidrule(l){2-3}
& \dots & \dots \\
\bottomrule
\end{tabularx}

\end{table}

\subsection{The Use of Large Language Models}
In this work, LLMs are used for three purposes: (1) we use GPT-4o and Claude Sonnet 4 to translate created datasets to the English version; (2) we use GPT-4o to evaluate whether model responses match the reference answers in Section \ref{sec:Cultural Awareness Grounding Validation} and Section \ref{sec:Cultural Knowledge Generalization}; and (3) we use GPT-4o to polish the manuscript.

\subsection{Data Sensitivity and Privacy Discussion}

In this section, we discuss data sensitivity and privacy for three modalities.

\paragraph{Text.} We collect culture-related textual facts from the public internet and filter out any content containing racism or hate speech. Our “knowledge points” are a few sentences manually summarized by annotators, and all questions are authored from scratch, so no infringement issues are involved.

\paragraph{Images.} We use only images under Creative Commons licenses and strictly for research purposes. We apply an automated face-blurring tool ~\citep{ren2023pbidr} to protect privacy, followed by manual review to catch any misses or false positives. Please note that faces of historical figures or cartoon characters are not masked. Our images do not contain pornography, violence, or other harmful content.

\paragraph{Speech.} We obtained voice samples from individuals outside the data team for speech synthesis, without disclosing any personal information. Consent was obtained prior to recording, and the audio is used solely for research.

\end{CJK}
\end{document}